\documentclass[final,5p,times,twocolumn]{elsarticle}

\usepackage{amsmath,amssymb,amsfonts}
\usepackage{algorithmic}
\usepackage{graphicx}
\usepackage{textcomp}
\usepackage[linesnumbered,ruled]{algorithm2e}
\usepackage{booktabs}
\usepackage{epsfig}
\usepackage{subfigure}
\usepackage{caption,setspace}
\usepackage{xcolor}
\usepackage{stfloats}

\let\vec\mathbf

\journal{Knowledge-Based Systems}

\begin{document}

\begin{frontmatter}



\title{An Improved LSHADE-RSP Algorithm with the Cauchy Perturbation: iLSHADE-RSP}


\author[affil1]{Tae Jong Choi}
\author[affil2]{Chang Wook Ahn}

\address[affil1]{Department of Artifical Intelligence and Software, Kyungil University\\ 50, Gamasil-gil, Hayang-eup, Gyeongsan-si, Gyeongsangbuk-do, Republic of Korea}
\address[affil2]{Artificial Intelligence Graduate School, Gwangju Institute of Science and Technology (GIST)\\ 123, Cheomdangwagi-ro, Buk-gu, Gwangju, Republic of Korea}

\begin{abstract}
A new method for improving the optimization performance of a state-of-the-art differential evolution (DE) variant is proposed in this paper. The technique can increase the exploration by adopting the long-tailed property of the Cauchy distribution, which helps the algorithm to generate a trial vector with great diversity. Compared to the previous approaches, the proposed approach perturbs a target vector instead of a mutant vector based on a jumping rate. We applied the proposed approach to LSHADE-RSP ranked second place in the CEC 2018 competition on single objective real-valued optimization. A set of 30 different and difficult optimization problems is used to evaluate the optimization performance of the improved LSHADE-RSP. Our experimental results verify that the improved LSHADE-RSP significantly outperformed not only its predecessor LSHADE-RSP but also several cutting-edge DE variants in terms of convergence speed and solution accuracy.
\end{abstract}


\end{frontmatter}


\section{Introduction}
\label{sec:Introduction}
A population-based metaheuristic optimization method called evolutionary algorithms (EAs) is designed based on Darwin's theory of natural selection. EAs generate a set of initial candidate solutions and update them iteratively with artificially designed evolutionary operators. As compared to traditional search algorithms, EAs are global, robust, and can be applied to any problem.

It is important to establish a balance between exploration and exploitation to improve the optimization performance of EAs. Matej et al. \cite{vcrepinvsek2013exploration} stated that ``Exploration is the process of visiting entirely new regions of a search space, whilst exploitation is the process of visiting those regions of a search space within the neighborhood of previously visited points." If an EA has too strong exploration, it might not be beneficial from existing candidate solutions \cite{eiben1998evolutionary}. On the other hand, if an EA has too strong exploitation, the probability of finding an optimal solution might be decreased \cite{eiben1998evolutionary}. Many researchers have investigated a number of approaches for balancing the two cornerstones \cite{vcrepinvsek2013exploration}.

Differential evolution (DE) proposed by Storn and Price \cite{storn1997differential, price2006differential} is one of the most successful EAs to deal with mathematical optimization. DE distributes its candidate solutions over the search boundaries of an optimization problem and updates them iteratively with vector difference based evolutionary operators. DE has two main advantages over other EAs: 1) it has a simple structure and a few control parameters, and 2) the effectiveness of DE has been demonstrated on various real-world problems \cite{das2010differential, das2016recent}. Among numerous DE variants, L-SHADE variants \cite{zhang2009jade, tanabe2013success, tanabe2014improving, brest2016shade, brest2017single, awad2016ensemble, awad2017novel, awad2017ensemble, awad2018ensemble, mohamed2017lshade, stanovov2018lshade, yeh2019modified} frequently perform very well on various optimization problems. LSHADE-RSP \cite{stanovov2018lshade} was recently proposed and ranked second place in the CEC 2018 competition on single objective real-valued optimization.

Although LSHADE-RSP has shown excellent performance, it has the following problem: LSHADE-RSP uses a rank-based selective pressure scheme, which increases the greediness and boosts the convergence speed. However, it may cause premature convergence in which all the candidate solutions fall into the local optimum of an optimization problem and cannot escape from there \cite{baker1985adaptive, de1975analysis, louis1993syntactic, goldberg1987finite}. Although LSHADE-RSP uses a setting for increasing the number of $p$best individuals in an effort to mitigate the problem, it may not be sufficient. In other words, LSHADE-RSP may fail to achieve exploration and exploitation.

In this paper, we proposed a new method for improving the optimization performance of LSHADE-RSP. The technique perturbs a target vector with the Cauchy distribution based on a jumping rate, which helps the algorithm to generate a trial vector with great diversity. Therefore, the technique can increase the probability of finding an optimal solution by adopting the long-tailed property of the Cauchy distribution. In the literature of DE, some researchers have demonstrated the effectiveness of using the Cauchy distribution in the phase of the recombination \cite{choi2020advanced, choi2019acm, zhang2017adaptive, ali2011improving, qin2010multi}. The novelty of the proposed approach lies in the perturbation of a target vector instead of a mutant vector. We named the combination of LSHADE-RSP and the proposed approach as iLSHADE-RSP.

We carried out experiments to evaluate the optimization performance of the proposed algorithm on the CEC 2017 test suite \cite{awad2016problem}. Our experimental results verify that the proposed algorithm significantly outperformed not only its predecessor LSHADE-RSP but also several state-of-the-art DE variants in terms of convergence speed and solution accuracy.

The rest of this paper is organized as follows: We introduce the background of this paper in Section \ref{sec:Background}. In Section \ref{sec:LiteratureReview}, we review the relevant literature to know, especially for L-SHADE variants. In Section \ref{sec:ProposedAlgorithm}, the details of the proposed algorithm is explained. We describe the experimental setup in Section \ref{sec:ExperimentalSetup}. We present the experimental results and discussion in Section \ref{sec:ExperimentalResultsAndDiscussion}. Finally, we conclude this paper in Section \ref{sec:Conclusion}.


\section{Background}
\label{sec:Background}

\subsection{Differential Evolution}
Since it was introduced, DE \cite{storn1997differential, price2006differential} has received much attention because of simplicity and applicability. At the beginning of an optimization process, DE generates a set of $NP$ initial candidate solutions as follows.

\begin{equation}
\vec{P}_{g} = (\vec{x}_{1,g}, \vec{x}_{2,g}, \cdots, \vec{x}_{NP,g})
\end{equation}

\noindent
where $\vec{P}_{g}$ denotes a population at generation $g$. Each candidate solution denoted by $\vec{x}_{i,g} = (x_{i,g}^{1}, x_{i,g}^{2}, \cdots, x_{i,g}^{D})$ is a $D$-dimensional vector. DE updates the candidate solutions iteratively with vector difference based evolutionary operators, such as mutation, crossover, and selection, to search for the global optimum of an optimization problem. The mutation and crossover operators create a set of $NP$ offspring, and the selection operator creates a population for the next generation by comparing the fitness value of a candidate solution and that of its corresponding offspring. DE returns the current best solution when it reaches the maximum number of generations $G_{max}$ or function evaluations $NFE_{max}$.

\subsubsection{Initialization}
At the beginning of an optimization process, DE distributes its candidate solutions over the search boundaries of an optimization problem with the initialization operator. Each candidate solution is initialized as follows.

\begin{equation}
x_{i,0}^{j} = x_{min}^{j} + rand_{i}^{j} \cdot (x_{max}^{j} - x_{min}^{j})
\end{equation}

\noindent
where $\vec{x}_{min} = (x_{min}^{1}, x_{min}^{2}, \cdots, x_{min}^{D})$ and $\vec{x}_{max} = (x_{max}^{1}, x_{max}^{2}, \cdots, x_{max}^{D})$ denote the lower and upper search boundaries of an optimization problem, respectively. Also, $rand_{i}^{j}$ denotes a uniformly distributed random number between $[0, 1]$.

\subsubsection{Mutation}
A mutant vector $\vec{v}_{i,g}$ is created in the mutation operator. The six frequently used classical mutation strategies are listed as follows.

\begin{itemize}
\item DE/rand/1:\\ $\vec{v}_{i,g} = \vec{x}_{r_{1},g} + F \cdot (\vec{x}_{r_{2},g} - \vec{x}_{r_{3},g})$
\item DE/rand/2:\\ $\vec{v}_{i,g} = \vec{x}_{r_{1},g} + F \cdot (\vec{x}_{r_{2},g} - \vec{x}_{r_{3},g}) + F \cdot (\vec{x}_{r_{4},g} - \vec{x}_{r_{5},g})$
\item DE/best/1:\\ $\vec{v}_{i,g} = \vec{x}_{best,g} + F \cdot (\vec{x}_{r_{1},g} - \vec{x}_{r_{2},g})$
\item DE/best/2:\\ $\vec{v}_{i,g} = \vec{x}_{best,g} + F \cdot (\vec{x}_{r_{1},g} - \vec{x}_{r_{2},g}) + F \cdot (\vec{x}_{r_{3},g} - \vec{x}_{r_{4},g})$
\item DE/current-to-best/1:\\ $\vec{v}_{i,g} = \vec{x}_{i,g} + F \cdot (\vec{x}_{best,g} - \vec{x}_{i,g}) + F \cdot (\vec{x}_{r_{1},g} - \vec{x}_{r_{2},g})$
\item DE/current-to-rand/1:\\ $\vec{v}_{i,g} = \vec{x}_{i,g} + K \cdot (\vec{x}_{r_{1},g} - \vec{x}_{i,g}) + F \cdot (\vec{x}_{r_{2},g} - \vec{x}_{r_{3},g})$
\end{itemize}

\noindent
where $r_{1}$, $r_{2}$, $r_{3}$, $r_{4}$, $r_{5}$ denote mutually different random indices within $\{1, 2, \cdots, NP\}$, which are also different from $i$. Moreover, $\vec{x}_{best,g}$ denotes the current best candidate solution. Furthermore, $F$ denotes a scaling factor, and $K$ denotes a uniformly distributed random number between $[0, 1]$.

\subsubsection{Crossover}
A trial vector $\vec{u}_{i,g}$ is created in the crossover operator. The binomial crossover frequently used creates a trial vector as follows.

\begin{equation}
u_{i,g}^{j} = \left\{ \begin{array}{ll}
v_{i,g}^{j} & \textrm{if $rand_{i}^{j} < CR$ or $j = j_{rand}$} \\
x_{i,g}^{j} & \textrm{otherwise}
\end{array} \right.
\end{equation}

\noindent
where $j_{rand}$ denotes a random index within $\{1, 2, \cdots, D\}$. Also, $CR$ denotes a crossover rate. The exponential crossover creates a trial vector as follows.

\begin{equation}
u_{i,g}^{j} = \left\{ \begin{array}{ll}
v_{i,g}^{j} & \textrm{if $j = \langle n \rangle_{D}, \langle n+1 \rangle_{D}, \cdots, \langle n+L-1 \rangle_{D}$} \\
x_{i,g}^{j} & \textrm{otherwise} 
\end{array} \right.
\end{equation}

\noindent
where $n$ denotes a random index within $\{1, 2, \cdots, D\}$, and $L$ denotes the number of elements, which can be calculated as follows.

$L = 0$

DO \{ $L = L + 1$ \}

WHILE ($(rand_{i}^{j} < CR)$ AND $(L < D)$)

\noindent
Also, $\langle \cdot \rangle_{D}$ denotes the modulo of $D$.

\subsubsection{Selection}
DE creates a population for the next generation with the selection operator. The selection operator compares the fitness value of a candidate solution ($\vec{x}_{i,g}$) and that of its corresponding offspring ($\vec{u}_{i,g}$) and picks the better one in terms of solution accuracy as follows.

\begin{equation}
\vec{x}_{i,g+1} = \left\{ \begin{array}{ll}
\vec{u}_{i,g} & \textrm{if $f(\vec{u}_{i,g}) \leq f(\vec{x}_{i,g})$} \\
\vec{x}_{i,g} & \textrm{otherwise}
\end{array} \right.
\end{equation}

\noindent
where $f(\vec{x})$ denotes an optimization problem to be minimized.

\subsection{Analysis of Cauchy Distribution}
The Cauchy distribution is a family of continuous probability distributions, which is stable and has a probability density function (PDF), which can be expressed analytically. As compared to the Gaussian distribution, the Cauchy distribution has a higher peak and a longer tail. The Cauchy distribution has two parameters: the location parameter $x_{0}$ and the scale parameter $\gamma$. The Cauchy distribution has a short and wide PDF if the scale parameter is high, while a tall and narrow PDF if the scale parameter is low. The PDF of the Cauchy distribution with $x_{0}$ and $\gamma$ can be defined as follows.

\begin{equation}
f(x; x_{0}, \gamma) = \frac{1}{\pi \gamma[1 + (\frac{x - x_{0}}{\gamma})^{2}]} = \frac{1}{\pi} \biggl[ \frac{\gamma}{(x - x_{0})^{2} + \gamma^{2}} \biggl]
\end{equation}

\noindent
Additionally, the cumulative distribution function of the Cauchy distribution with $x_{0}$ and $\gamma$ can be defined as follows.

\begin{equation}
F(x; x_{0}, \gamma) = \frac{1}{\pi} arctan \biggl( \frac{x - x_{0}}{\gamma} \biggl) + \frac{1}{2}
\end{equation}

\begin{figure}[!t]
\centering
\includegraphics[width=\linewidth]{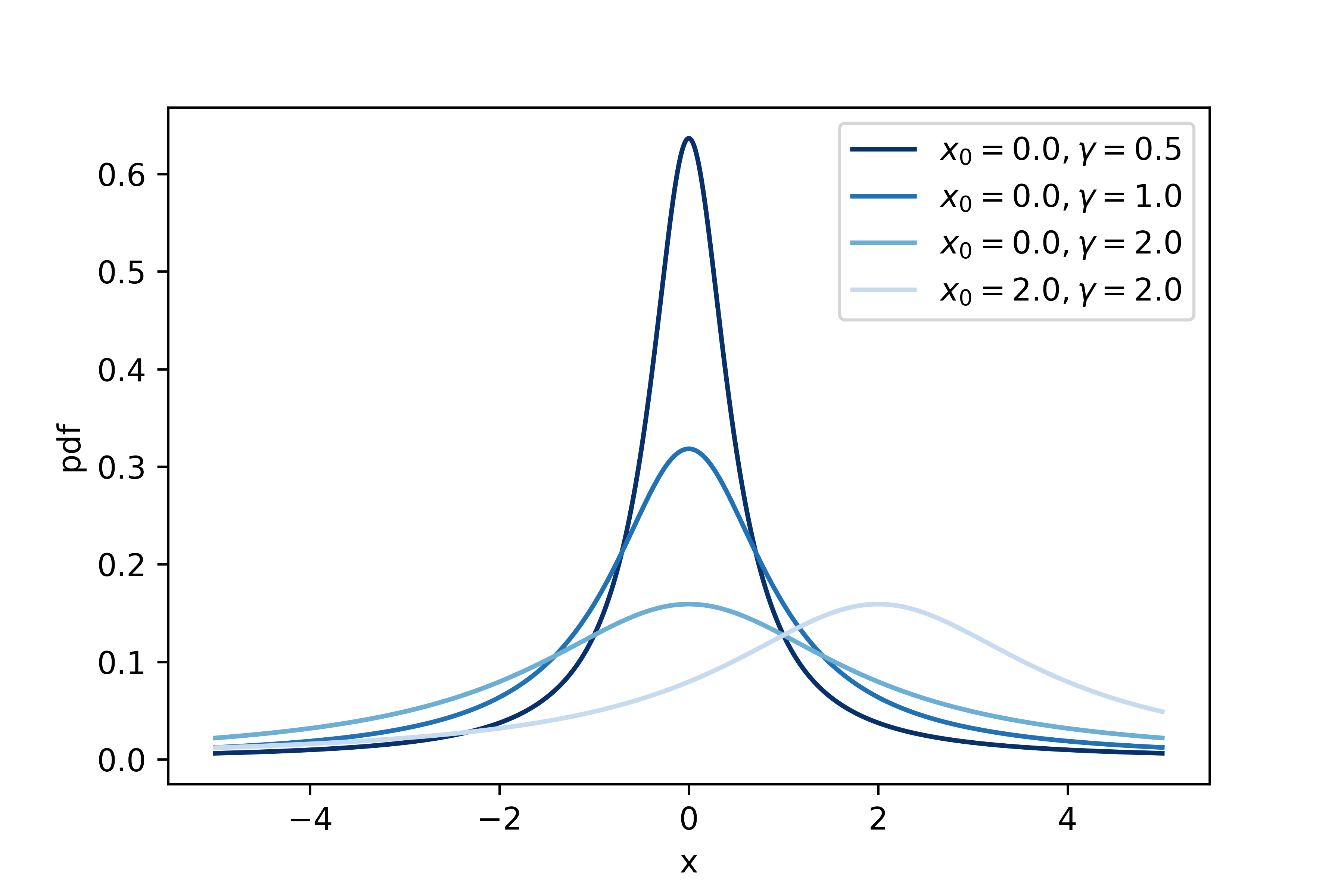}
\caption{The four different PDFs of the Cauchy distribution}
\label{fig_pdfs_cauchy}
\end{figure}

\noindent
Fig. \ref{fig_pdfs_cauchy} shows the four different PDFs of the Cauchy distribution.

\section{Literature Review}
\label{sec:LiteratureReview}
Since it was introduced, many researchers have developed new methods for DE, such as

\begin{itemize}
\item Adaptive parameters with single mutation strategy DE algorithms \cite{brest2006self, brest2008population, choi2013adaptive, leon2016adapting, choi2017improved, choi2017adaptive2, bujok2017enhanced, choi2018asynchronous2, choi2018performance},
\item Adaptive parameters with multiple mutation strategy DE algorithms \cite{qin2008differential, mallipeddi2011differential, wang2011differential, choi2014adaptive1, choi2014adaptive2, choi2015adaptive1, choi2015adaptive2, wu2016differential, wu2018ensemble, choi2017adaptive1, choi2018asynchronous1},
\item Hybrid DE algorithms \cite{rahnamayan2008opposition, wang2011enhanced, park2015stochastic, choi2019fast, qin2010multi, ali2011improving, zhang2017adaptive, choi2019acm, choi2019adaptive, choi2020advanced},
\item DE algorithms with sampling explicit probabilistic models \cite{mininno2010compact, zhong2012sde, wang2013gaussian}.
\end{itemize}

\noindent
For more detailed information, please refer to the following papers \cite{das2010differential, das2016recent, al2018algorithmic, opara2019differential}.

Among numerous DE variants, L-SHADE variants \cite{zhang2009jade, tanabe2013success, tanabe2014improving, brest2016shade, brest2017single, awad2016ensemble, awad2017novel, awad2017ensemble, awad2018ensemble, mohamed2017lshade, stanovov2018lshade, yeh2019modified} frequently perform very well on various optimization problems. JADE \cite{zhang2009jade} proposed by Zhang and Sanderson is considered to be the origin of L-SHADE variants. JADE uses a new mutation strategy DE/current-to-$p$best/1 with an external archive. The new mutation strategy is the generalization of a classical mutation strategy DE/current-to-best/1, and the external archive supplies the progress of an evolutionary process. These modifications improve the exploration of the algorithm. JADE also uses a learning process-based adaptive parameter control to adjust the control parameters $F$ and $CR$ automatically. In the experiments, JADE outperformed several algorithms, including jDE \cite{brest2006self}, SaDE \cite{qin2008differential}, and PSO \cite{trelea2003particle}.

Tanabe and Fukunaga proposed SHADE \cite{tanabe2013success}, an enhancement to JADE. The main difference between SHADE and JADE is that SHADE utilizes historical memories, which store the average of successfully evolved individuals' control parameters $F$ and $CR$ to adjust the control parameters $F$ and $CR$ automatically. In the experiments, SHADE outperformed several algorithms, including CoDE \cite{wang2011differential}, EPSDE \cite{mallipeddi2011differential}, and dynNP-DE \cite{brest2008population}. Tanabe and Fukunaga later proposed L-SHADE \cite{tanabe2014improving}, an enhancement to SHADE. The main difference between L-SHADE and SHADE is that L-SHADE utilizes linear population size reduction (LPSR), which gradually reduces the population size as a linear function to establish a balance between exploration and exploitation. In the experiments, L-SHADE outperformed several algorithms, including NBIPOP-$_{\text{A}}$CMA-ES \cite{liao2013benchmark} and iCMAES-ILS \cite{lacroix2013dynamically}.

Brest et al. proposed an improved version of L-SHADE called iL-SHADE \cite{brest2016shade}. iL-SHADE updates historical memories $\vec{\mu}_{F}$ and $\vec{\mu}_{CR}$ by calculating the average of old and new values. iL-SHADE also gradually reduces the $p$ value of DE/current-to-$p$best/1 as a linear function. iL-SHADE was ranked third place in the CEC 2016 competition on single objective real-valued optimization. Brest et al. later proposed an improved version of iL-SHADE called jSO \cite{brest2017single}. jSO uses a new mutation strategy DE/current-to-$p$best-w/1, which assigns a lower scaling factor $F_{w}$ at the early stage of an evolutionary process and a higher scaling factor $F_{w}$ at the late stage of an evolutionary process. jSO was ranked second place in the CEC 2017 competitions on single objective real-valued optimization.

Awad et al. proposed LSHADE-EpSin \cite{awad2016ensemble} based on L-SHADE, which utilizes a new ensemble sinusoidal approach for tuning the scaling factor $F$ in an adaptive manner. The new ensemble sinusoidal approach is the combination of two sinusoidal waves whose objective is to establish a balance between exploration and exploitation. LSHADE-EpSin also utilizes a random walk at the late stage of an evolutionary process. Awad et al. proposed L-$conv$SHADE \cite{awad2017novel} based on L-SHADE, which utilizes a new crossover operator based on covariance matrix adaptation with Euclidean neighborhood for rotation invariance. Awad et al. later proposed LSHADE-$cn$EpSin \cite{awad2017ensemble}, which is the combination of LSHADE-EpSin and L-$conv$SHADE with two major modifications: 1) a new ensemble sinusoidal approach with a learning process-based adaptive parameter control and 2) a new crossover operator based on covariance matrix adaptation with Euclidean neighborhood for rotation invariance. Finally, Awad et al. proposed EsDE$_{\text{r}}$-NR \cite{awad2018ensemble} based on LSHADE-EpSin, which gradually reduces the population size with a niching-based approach.

Mohamed et al. proposed LSHADE-SPACMA \cite{mohamed2017lshade}, an enhancement to L-SHADE, which is a hybrid algorithm between LSHADE-SPA \cite{mohamed2017lshade} and a modified version of CMA-ES \cite{hansen2006cma}. LSHADE-SPA uses a new semi-parameter control for tuning the scaling factor $F$ in an adaptive manner. The modified version of CMA-ES undergoes the phase of the crossover, which improves the exploration of the algorithm.

Yet et al. propose mL-SHADE \cite{yeh2019modified}, an enhancement to L-SHADE, in which three major modifications are made: 1) a terminal value for the control parameter $CR$ is excluded, 2) a polynomial mutation strategy is included, and 3) a perturbation for historical memories is included. These modifications improve the exploration of the algorithm.


\section{Proposed Algorithm}
\label{sec:ProposedAlgorithm}
This section describes an improved LSHADE-RSP called iLSHADE-RSP, which employs a modified recombination operator, which calculates a perturbation of a target vector with the Cauchy distribution.

\subsection{Mutation Strategy}
The proposed algorithm uses a new mutation strategy DE/current-to-$p$best/r \cite{stanovov2018lshade}. The strategy is designed based on a rank-based selective pressure scheme \cite{gong2013differential}, which proportionally selects two donor vectors $\vec{x}_{r_{1},g}$ and $\vec{x}_{r_{2},g}$ with respect to the fitness value. The higher the ranking of a candidate solution has, the more opportunity it will be selected. The strategy can be defined as follows.

\begin{itemize}
\item DE/current-to-$p$best/r:\\ $\vec{v_{i,g}} = \vec{x_{i,g}} + F_{w} \cdot (\vec{x_{pbest,g}} - \vec{x_{i,g}}) + F \cdot (\vec{x_{pr_{1},g}} - \vec{\tilde{x}_{pr_{2},g}})$
\end{itemize}

\noindent
where
\begin{equation*}
F_{w} = \left\{ \begin{array}{ll}
0.7 \cdot F & \textrm{if $0 \leq NFE < 0.2 \cdot NFE_{max}$} \\
0.8 \cdot F & \textrm{if $0.2 \cdot NFE_{max} \leq NFE < 0.4 \cdot NFE_{max}$} \\
1.2 \cdot F & \textrm{otherwise}
\end{array} \right.
\end{equation*}

\noindent
where $\vec{x_{pbest,g}}$ denotes one of the top $100p\%$ individuals with $p \in (0, 1]$. Also, $\vec{x_{pr_{1},g}}$ denotes a random donor vector from a population based on rank-based probabilities, and $\vec{\tilde{x}_{pr_{2},g}}$ denotes a random donor vector from a population based on rank-based probabilities or from an external archive. The probability of the $i$th individual being selected can be calculated as follows.

\begin{equation}
pr_{i} = \frac{Rank_{i}}{\sum_{j=1}^{NP_{g}}(Rank_{j})}
\end{equation}

\noindent
where

\begin{equation}
Rank_{i} = k \cdot (NP_{g} - i) + 1
\end{equation}

\noindent
where $k$ denotes a rank greediness factor. Additionally, LSHADE-RSP uses a setting for increasing the number of $p$best individuals, which can be calculated as follows.

\begin{equation}
p = 0.085 \cdot \Big( 1 + \frac{NFE}{NFE_{max}} \Big)
\end{equation}

\subsection{Linear Population Size Reduction}
The proposed algorithm uses linear population size reduction (LPSR) \cite{tanabe2014improving} to establish a balance between exploration and exploitation. The idea behind LPSR is to use a higher population size at the beginning of an optimization process and gradually reduce it as a linear function. At the end of each generation, LPSR calculates the population size for the next generation $NP_{g+1}$ as follows.

\begin{equation}
NP_{g+1} = round \Big[ NP_{init} - \frac{NES}{NES_{max}} \cdot \Big( NP_{init} - NP_{fin} \Big) \Big]
\end{equation}

\noindent
where $NP_{init}$ and $NP_{fin}$ denote the initial and final population sizes, respectively. If the next population size $NP_{g+1}$ is smaller than the current one $NP_{g}$, the worst $NP_{g} - NP_{g+1}$ candidate solutions with respect to the fitness value are discarded. For the initial and final population sizes, LSHADE-RSP uses $NP_{init} = round \Big( sqrt(D) * log(D) * 25) \Big)$ and $NP_{fin} = 4$.

\subsubsection{Adaptive Parameter Control}
The proposed algorithm uses a learning process-based adaptive parameter control \cite{brest2017single} to adjust the control parameters $F$ and $CR$ automatically. Each candidate solution has its control parameters $F_{i,g}$ and $CR_{i,g}$. At each generation, the control parameters $F_{i,g}$ and $CR_{i,g}$ are calculated as follows.

\begin{equation}
F_{i,g} = rndc_{i}( M_{F,r}, 0.1 )
\end{equation}

\begin{equation}
CR_{i,g} = rndn_{i}( M_{CR,r}, 0.1 )
\end{equation}

\noindent
where $rndc_{i}$ and $rndn_{i}$ denote the Cauchy and Gaussian distributions, respectively. Also, $M_{F,r}$ and $M_{CR,r}$ denote randomly selected values from historical memories $\vec{\mu}_{F}$ and $\vec{\mu}_{CR}$, respectively. The scaling factor is recalculated if $F_{i,g} \leq 0$ or truncated to 1 if $F_{i,g} > 1$. The crossover rate is first truncated to $[0, 1]$. After that, the crossover rate is modified as follows.

\begin{equation*}
CR_{i,g} = \left\{ \begin{array}{ll}
0.7 & \textrm{if $CR_{i} < 0.7$ and $NFE < 0.25 \cdot NFE_{max}$} \\
0.6 & \textrm{if $CR_{i} < 0.6$ and $NFE < 0.5 \cdot NFE_{max}$} \\
CR_{i,g} & \textrm{otherwise}
\end{array} \right.
\end{equation*}

The historical memories $\vec{\mu}_{F}$ and $\vec{\mu}_{CR}$ store the successfully evolved candidate solutions' control parameters. The capacity of the historical memories is $H$. At the beginning of an optimization process, all the entries, except the last one, of the memory $\vec{\mu}_{F}$ are initialized to 0.3. Similarly, all the entries, except the last one, of the memory $\vec{\mu}_{CR}$ are initialized to 0.8. The last entry of the memories $\vec{\mu}_{F}$ and $\vec{\mu}_{CR}$ always keep 0.9 during the optimization process. After the selection operator, the successfully evolved candidate solutions' control parameters are stored in $S_{F}$ and $S_{CR}$. Then, one of the entries of the memories is updated as follows.

\begin{equation}
M_{F,k} = \left\{ \begin{array}{ll}
mean_{WL}(S_{F}) & \textrm{if $S_{F} \neq \emptyset$} \\
M_{F,k} & \textrm{otherwise} 
\end{array} \right.
\end{equation}

\begin{equation}
M_{CR,k} = \left\{ \begin{array}{ll}
mean_{WL}(S_{CR}) & \textrm{if $S_{CR} \neq \emptyset$} \\
M_{CR,k} & \textrm{otherwise} 
\end{array} \right.
\end{equation}

\noindent
where $meanw_{WL}$ denotes the weighted Lehmer mean, which takes into consideration the improvement in fitness values between candidate solutions and their corresponding offspring.


\subsection{Cauchy Perturbation}
LSHADE-RSP \cite{stanovov2018lshade} uses a rank-based selective pressure scheme, which tends to select higher ranking candidate solutions as donor vectors. Therefore, the scheme can increase the greediness, which can boost the convergence speed. However, the scheme may decrease the solution accuracy because of premature convergence in which all the candidate solutions fall into the local optimum of an optimization problem and cannot escape from there \cite{baker1985adaptive, de1975analysis, louis1993syntactic, goldberg1987finite}. Although LSHADE-RSP uses a setting for increasing the number of $p$best individuals, it may not be sufficient to compensate for the increased greediness.

To improve the exploration property of EAs, many researchers have developed new methods. Among them, using a long-tailed stable distribution, such as the Cauchy or L\'evy distribution, in the phase of the recombination is one of the popular ones. By using a long-tailed stable distribution, EAs can generate candidate solutions over large distances, which can improve the exploration property. In the literature of DE, some researchers have demonstrated the effectiveness of using a long-tailed stable distribution in the phase of the recombination \cite{ali2011improving, qin2010multi, zhang2017adaptive}. This motivated us to devise a modified recombination operator for LSHADE-RSP.

\begin{figure}[!t]
 \centering
 \subfigure[Original DE/rand/1/bin]{
  \includegraphics[scale=0.9]{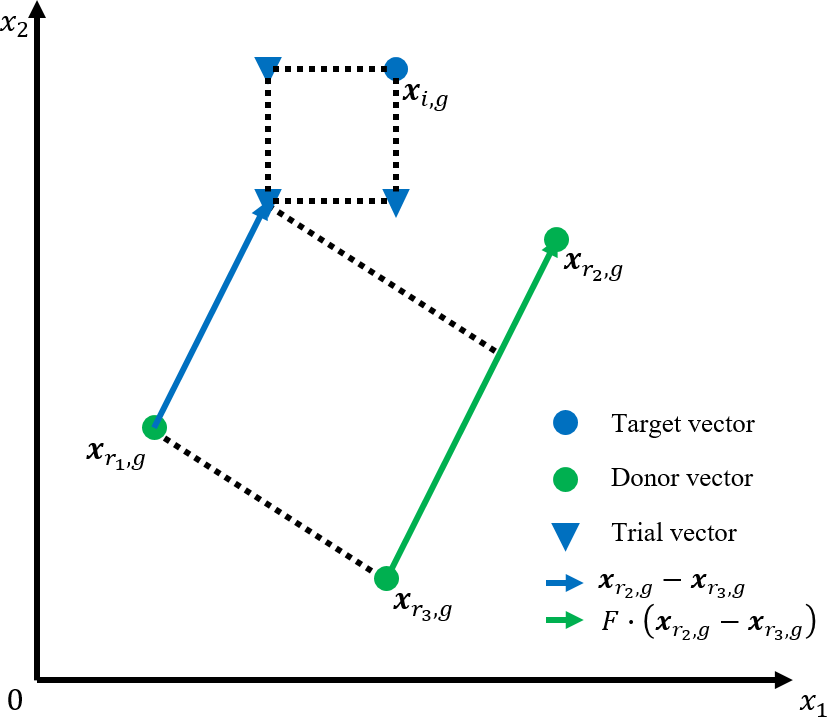}
   }
 \subfigure[Modified DE/rand/1/bin]{
  \includegraphics[scale=0.9]{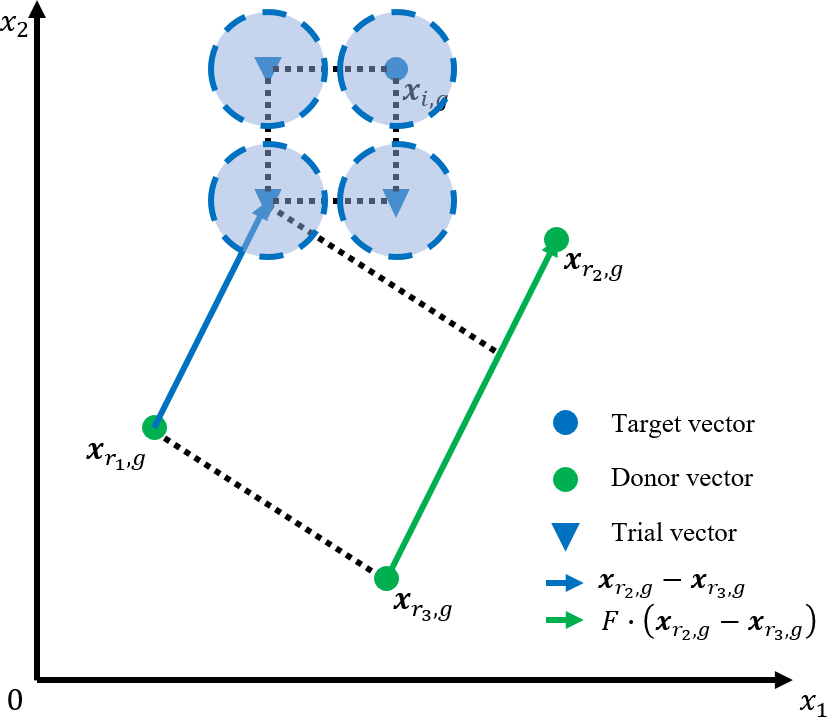}
   }
 \caption[]{This figure illustrates the behavior of the modified recombination operator with DE/rand/1/bin. For simplicity of explanation, we chose DE/rand/1/bin instead of DE/current-to-$p$best/r. As we can see from the figure, the modified DE/rand/1/bin can explore larger feasible regions than the original DE/rand/1/bin. This is because the modified DE/rand/1/bin perturbs a target vector, which increases the number of possible locations for its corresponding trial vector significantly.}
 \label{fig:CauchyPerturbation}
\end{figure}

The idea behind the modified recombination operator is simple. When creating a trial vector, the operator first perturbs a target vector with the Cauchy distribution. After that, the operator creates a trial vector by recombining the perturbed target vector and its corresponding mutant vector. Therefore, a much different trial vector can be created by adopting the long-tail property of the Cauchy distribution. The novelty of the proposed approach lies in the perturbation of a target vector instead of a mutant vector. Fig. \ref{fig:CauchyPerturbation} illustrates the behavior of the modified recombination operator with DE/rand/1/bin.

As we mentioned earlier, the modified recombination operator is an extension of the recombination operator of LSHADE-RSP. The original operator can be defined as follows.

\begin{equation}
u_{i,g}^{j} = \left\{ \begin{array}{ll}
x_{i,g}^{j} + F_{w} \cdot (x_{pbest,g}^{j} - x_{i,g}^{j}) & \textrm{if $rand_{i}^{j} < CR$ or $j = j_{rand}$} \\
\,\,\,\,\,\,\,\,\, + F \cdot (x_{pr_{1},g}^{j} - \tilde{x}_{pr_{2},g}^{j}) & \\
x_{i,g}^{j} & \textrm{otherwise}
\end{array} \right.
\label{originalRecombination}
\end{equation}

\noindent
where $x_{pbest,g}^{j}$ denotes the $j$th component of one of the top $100p\%$ individuals with $p \in (0, 1]$. Also, $x_{pr_{1},g}^{j}$ denotes the $j$th component of a random donor vector from a population based on rank-based probabilities, and $\tilde{x}_{pr_{2},g}^{j}$ denotes the $j$th component of a random donor vector from a population based on rank-based probabilities or from an external archive. The modified operator can be defined as follows.

\begin{equation}
u_{i,g}^{j} = \left\{ \begin{array}{ll}
x_{i,g}^{j} + F_{w} \cdot (x_{pbest,g}^{j} - x_{i,g}^{j}) & \textrm{if $rand_{i}^{j} < CR$ or $j = j_{rand}$} \\
\,\,\,\,\,\,\,\,\, + F \cdot (x_{pr_{1},g}^{j} - \tilde{x}_{pr_{2},g}^{j}) & \\
rndc_{i}^{j}(x_{i,g}^{j}, 0.1) & \textrm{otherwise}
\end{array} \right.
\label{modifiedRecombination}
\end{equation}

\noindent
where $rndc_{i}^{j}$ denotes the Cauchy distribution.

The proposed algorithm alternately applies one of the two recombination operators according to the jumping rate $p_{j}$. When creating a trial vector, the proposed algorithm applies the original operator if a random number is higher than or equal to the rate. Otherwise, the proposed algorithm applies the modified operator. Algorithm \ref{alg:iLSHADE-RSP} shows the pseudo-code of the proposed algorithm.

\begin{algorithm}[h!]
    \SetKwInOut{Input}{Input}
    \SetKwInOut{Output}{Output}

    \Input{Objective function $f(\vec{x})$, lower bound $\vec{x_{min}}$, upper bound $\vec{x_{max}}$, maximum number of function evaluations $NFE_{max}$, and jumping rate $p_{j}$}
    \Output{Final best objective value $f(\vec{x_{best,G_{max}}})$}
    \tcc{Initialization}
    Set function evaluation $NFE \leftarrow 0$\;
    Set generation $g \leftarrow 1$\;
    Initialize population $\vec{P}_{g} = (\vec{x}_{1,g}, \cdots, \vec{x}_{NP,g})$ randomly\;
    Set archive $\vec{A} \leftarrow \emptyset$\;
    Set all elements in $\vec{\mu}_{F}$ to 0.3\;
    Set all elements in $\vec{\mu}_{CR}$ to 0.8\;
    \tcc{Iteration}
    \While{None of termination criteria is satisfied}{
        \tcc{Recombination operator}
        Set $S_{F} \leftarrow \emptyset$, $S_{CR} \leftarrow \emptyset$\;
        \For{$i = 0;\ i < NP;\ i = i + 1$}{
            Assign $F_{i,g}$, $CR_{i,g}$ using Algorithm \ref{alg:parameterAssignment}\;
            \eIf{$rand_{i} \leq p_{j}$}
            {
                $\vec{u}_{i,g} \leftarrow$ the modified operator using Eq. (\ref{modifiedRecombination})\;
            }
            {
                $\vec{u}_{i,g} \leftarrow$ the original operator using Eq. (\ref{originalRecombination})\;
            }
        }
        \tcc{Selection operator}
        \For{$i = 0;\ i < NP;\ i = i + 1$}{
            \eIf{$f(\vec{u_{i,g}}) \leq f(\vec{x_{i,g}})$}
            {
                $\vec{x_{i,g+1}} \leftarrow \vec{u_{i,g}}$\;
                $\vec{x_{i,g}} \rightarrow \vec{A}$\;
                $F_{i,g} \rightarrow S_{F}$, $CR_{i,g} \rightarrow S_{CR}$\;
            }
            {
                $\vec{x_{i,g+1}} \leftarrow \vec{x_{i,g}}$\;
            }
        }
        Shrink $\vec{P}_{g+1}$ by discarding worst solutions\;
        Shrink $\vec{A}$ by discarding random solutions\;
        Update $\vec{\mu}_{F}$, $\vec{\mu}_{CR}$\;
        $p \leftarrow 0.085 \cdot \Big( 1 + \frac{NFE}{NFE_{max}} \Big)$\;
        $g \leftarrow g + 1$\;
    }
    \caption{iLSHADE-RSP}
    \label{alg:iLSHADE-RSP}
\end{algorithm}

\begin{algorithm}[h!]
    Select $r_{i}$ from $[1, H]$ randomly\;
    \If{$r_{i} = H$}
    {
        $M_{F,r_{i}} \leftarrow 0.9$\;
        $M_{CR,r_{i}} \leftarrow 0.9$\;
    }
    $F_{i,g} \leftarrow rndc_{i}(M_{F,r_{i}}, 0.1)$\;
    \If{$g < 0.6 \cdot NFE_{max}$ \textbf{ and } $F_{i,g} > 0.7$}
    {
        $F_{i,g} \leftarrow 0.7$
    }
    \eIf{$M_{CR,r_{i}} < 0$}
    {
        $CR_{i,g} \leftarrow 0$\;
    }
    {
        $CR_{i,g} \leftarrow rndn_{i}(M_{CR,r_{i}}, 0.1)$\;
    }
    \eIf{$g < 0.25 \cdot NFE_{max}$}
    {
        $CR_{i,g} \leftarrow max(CR_{i,g}, 0.7)$\;
    }
    {
        \If{$g < 0.5 \cdot NFE_{max}$}
        {
            $CR_{i,g} \leftarrow max(CR_{i,g}, 0.6)$\;
        }
    }
    \caption{Parameter Assignment}
    \label{alg:parameterAssignment}
\end{algorithm}


\section{Experimental Setup}
\label{sec:ExperimentalSetup}

\subsection{System Configuration}
All the following experiments were performed on Windows 10 Pro 64 bit of a PC with AMD Ryzen Threadripper 2990WX @ 3.0GHz. The proposed and comparison algorithms were developed in the C++ programming language with Visual Studio 2019 64 bit.

\subsection{Test Algorithms}
We used the following test algorithms for the comparative analysis.

\begin{itemize}
    \item iLSAHDE-RSP: the proposed algorithm.
    \item LSHADE-RSP \cite{stanovov2018lshade}: ranked the second place in the CEC 2018 competition on single objective optimization.
    \item jSO \cite{brest2017single}: ranked the second place in the CEC 2017 competition on single objective optimization.
    \item L-SHADE \cite{tanabe2014improving}: ranked the first place in the CEC 2014 competition on single objective optimization.
    \item SHADE \cite{tanabe2013success}: ranked the fourth place in the CEC 2013 competition on single objective optimization.
    \item JADE \cite{zhang2009jade}: the origin of L-SHADE variants.
    \item EDEV \cite{wu2018ensemble}: a multi-population-based DE variant.
    \item MPEDE \cite{wu2016differential}: a multi-population-based DE variant.
    \item CoDE \cite{wang2011differential}: a composite DE variant.
    \item EPSDE \cite{mallipeddi2011differential}: an ensemble DE variant.
    \item SaDE \cite{qin2008differential}: a self-adaptive DE variant.
    \item dynNP-DE \cite{brest2008population}: a self-adaptive DE variant.
\end{itemize}

The algorithms are six L-SHADE variants, two multi-population-based DE variants, and four well-known classical DE variants. The proposed algorithm introduces the jumping rate $p_{j}$. As shown in Section \ref{sec:ExperimentalResultsAndDiscussion}, the proposed algorithm works best with $p_{j} \in [0.15, 0.35]$. The proposed algorithm uses $p_{j} = 0.2$ in all the following experiments. Except for the jumping rate, the proposed algorithm uses the same values for the control parameters as its predecessor LSHADE-RSP.

\subsection{Test Functions}
To compare the proposed and comparison algorithms experimentally, we carried out experiments on the CEC 2017 test suite \cite{awad2016problem} in 10, 30, 50 and 100 dimensions. The CEC 2017 test suite has 30 different and difficult optimization problems, such as three unimodal test functions ($F_{1}$-$F_{3}$), seven simple multimodal test functions ($F_{4}$-$F_{10}$), ten expanded multimodal test functions ($F_{11}$-$F_{20}$), and ten hybrid composition test functions ($F_{21}$-$F_{30}$). A function is said to be unimodal if it has no local optima, while a function is said to be multimodal if it has multiple local optima.

According to the experimental setups of the test suite, the maximum number of function evaluations $NFE_{max}$ was set to $10,000 \cdot D$. Moreover, the search boundaries of the test suite were set to $[-100, 100]^{D}$. Furthermore, all the experimental results were obtained by 51 runs independently. For more detailed information, please refer to the following papers \cite{awad2016problem}.

\subsection{Performance Metrics}

\subsubsection{Function Error Value}
The function error value (FEV) is utilized to assess the test algorithm's accuracy. The FEV is the difference between the final best objective value of a test algorithm and the global optimum of an optimization problem, which can be defined as follows.

\begin{equation}
\textrm{FEV} = f(\vec{x}_{best,G_{max}}) - f(\vec{x}_{*})
\end{equation}

\noindent
where $f(\vec{x})$ denotes an objective function. Also, $\vec{x}_{best,G_{max}}$ and $\vec{x}_{*}$ denote the final best objective value and the global optimum, respectively.

\subsubsection{Statistical Test}
We utilized the Wilcoxon rank-sum test and the Friedman test with Hochberg's post hoc for the comparative analysis. The former is used to test the statistical significance of two test algorithms, while the latter is used to test the statistical significance of multiple test algorithms \cite{demvsar2006statistical}.


\section{Experimental Results and Discussion}
\label{sec:ExperimentalResultsAndDiscussion}



\begin{table*}[h!]
  \tiny
  \centering
  \caption{Means and standard deviations of FEVs of test algorithms on CEC 2017 test suite in 10 dimension}
\begin{tabular}{cccccccc}
    \toprule
          & iLSHADE-RSP & LSHADE-RSP & jSO   & L-SHADE & SHADE & JADE  & EDEV \\
          & MEAN (STD DEV) & MEAN (STD DEV) & MEAN (STD DEV) & MEAN (STD DEV) & MEAN (STD DEV) & MEAN (STD DEV) & MEAN (STD DEV) \\
    \midrule
    F1    & 0.00E+00 (0.00E+00) & 0.00E+00 (0.00E+00) = & 0.00E+00 (0.00E+00) = & 0.00E+00 (0.00E+00) = & 0.00E+00 (0.00E+00) = & 0.00E+00 (0.00E+00) = & 0.00E+00 (0.00E+00) = \\
    F2    & 0.00E+00 (0.00E+00) & 0.00E+00 (0.00E+00) = & 0.00E+00 (0.00E+00) = & 0.00E+00 (0.00E+00) = & 0.00E+00 (0.00E+00) = & 0.00E+00 (0.00E+00) = & 1.23E-09 (4.12E-09) = \\
    F3    & 0.00E+00 (0.00E+00) & 0.00E+00 (0.00E+00) = & 0.00E+00 (0.00E+00) = & 0.00E+00 (0.00E+00) = & 0.00E+00 (0.00E+00) = & 0.00E+00 (0.00E+00) = & 0.00E+00 (0.00E+00) = \\
    F4    & 0.00E+00 (0.00E+00) & 0.00E+00 (0.00E+00) = & 0.00E+00 (0.00E+00) = & 0.00E+00 (0.00E+00) = & 0.00E+00 (0.00E+00) = & 0.00E+00 (0.00E+00) = & 0.00E+00 (0.00E+00) = \\
    F5    & 1.29E+00 (8.03E-01) & 1.29E+00 (9.39E-01) = & 1.83E+00 (8.74E-01) - & 2.46E+00 (9.21E-01) - & 2.64E+00 (7.38E-01) - & 3.48E+00 (8.39E-01) - & 4.68E+00 (9.38E-01) - \\
    F6    & 2.91E-14 (5.02E-14) & 1.56E-14 (3.96E-14) = & 0.00E+00 (0.00E+00) + & 0.00E+00 (0.00E+00) + & 2.24E-15 (1.60E-14) + & 8.94E-14 (4.74E-14) - & 2.89E-12 (8.26E-12) - \\
\bottomrule
    \end{tabular}%
  \label{tab:cec17_10_wilcoxon}%
\footnotesize
\\The symbols ``+/=/-" indicate that the corresponding algorithm performed significantly better ($+$), not significantly better or worse ($=$), or significantly worse ($-$) compared to iLSHADE-RSP using the Wilcoxon rank-sum test with $\alpha = 0.05$ significance level.
\end{table*}%

\begin{table*}[h!]
  \tiny
  \centering
  \caption{Friedman test with Hochberg's post hoc for test algorithms on CEC 2017 test suite in 10 dimension}
    \begin{tabular}{cccccccccc}
    \toprule
          & Algorithm & Average ranking & z-value & p-value & Adj. p-value (Hochberg) & Sig.  &       & Test statistics &  \\
    \midrule
    1     & iLSHADE-RSP & 4.15  &       &       &       &       &       &       &  \\
    \cmidrule(l){2-3}
    2     & LSHADE-RSP & 4.42  & -2.86.E-01 & 7.75.E-01 & 7.75.E-01 & No    &       & N     & 30  \\
    3     & jSO   & 5.30  & -1.24.E+00 & 2.17.E-01 & 8.67.E-01 & No    &       & Chi-Square & 76.90  \\
    4     & L-SHADE & 5.23  & -1.16.E+00 & 2.45.E-01 & 7.34.E-01 & No    &       & df    & 11  \\
    5     & SHADE & 6.12  & -2.11.E+00 & 3.46.E-02 & 1.73.E-01 & No    &       & p-value & 5.85.E-12 \\
    6     & JADE  & 7.05  & -3.12.E+00 & 1.84.E-03 & 1.10.E-02 & Yes   &       & Sig.  & Yes \\
    7     & EDEV  & 7.35  & -3.44.E+00 & 5.87.E-04 & 4.11.E-03 & Yes   &       &       &  \\
    8     & MPEDE & 9.97  & -6.25.E+00 & 4.15.E-10 & 4.57.E-09 & Yes   &       &       &  \\
    9     & CoDE  & 7.87  & -3.99.E+00 & 6.54.E-05 & 5.89.E-04 & Yes   &       &       &  \\
    10    & EPSDE & 7.62  & -3.72.E+00 & 1.96.E-04 & 1.57.E-03 & Yes   &       &       &  \\
    11    & SaDE  & 7.90  & -4.03.E+00 & 5.62.E-05 & 5.62.E-04 & Yes   &       &       &  \\
    12    & dynNP-DE & 5.03  & -9.49.E-01 & 3.43.E-01 & 6.85.E-01 & No    &       &       &  \\
    \bottomrule
    \end{tabular}%
  \label{tab:cec17_10_friedman}%
\end{table*}%

\begin{table*}[h!]
  \tiny
  \centering
  \caption{Means and standard deviations of FEVs of test algorithms on CEC 2017 test suite in 30 dimension}
    \begin{tabular}{cccccccc}
    \toprule
          & iLSHADE-RSP & LSHADE-RSP & jSO   & L-SHADE & SHADE & JADE  & EDEV \\
          & MEAN (STD DEV) & MEAN (STD DEV) & MEAN (STD DEV) & MEAN (STD DEV) & MEAN (STD DEV) & MEAN (STD DEV) & MEAN (STD DEV) \\
    \midrule
    F1    & 1.67E-15 (4.62E-15) & 8.35E-16 (3.37E-15) = & 2.78E-16 (1.99E-15) = & 2.78E-16 (1.99E-15) = & 1.36E-14 (3.98E-15) - & 1.42E-14 (1.27E-29) - & 2.78E-16 (1.99E-15) = \\
    F2    & 0.00E+00 (0.00E+00) & 0.00E+00 (0.00E+00) = & 0.00E+00 (0.00E+00) = & 2.78E-15 (8.53E-15) = & 5.59E-13 (1.01E-12) - & 3.61E-13 (5.66E-13) - & 1.05E+13 (6.16E+13) - \\
    F3    & 2.56E-14 (2.85E-14) & 1.89E-14 (2.70E-14) = & 7.80E-15 (1.97E-14) + & 5.57E-15 (1.71E-14) + & 7.25E-14 (2.58E-14) - & 9.43E+03 (1.54E+04) - & 4.98E+03 (1.28E+04) = \\
    F4    & 2.27E+01 (7.02E-01) & 5.86E+01 (5.74E-14) - & 5.86E+01 (5.74E-14) - & 5.87E+01 (7.70E-01) - & 3.62E+01 (2.98E+01) = & 4.88E+01 (2.39E+01) - & 5.16E+01 (2.09E+01) - \\
    F5    & 7.89E+00 (2.38E+00) & 7.11E+00 (2.09E+00) = & 8.85E+00 (1.91E+00) = & 6.77E+00 (1.60E+00) = & 1.91E+01 (3.29E+00) - & 2.63E+01 (4.00E+00) - & 3.41E+01 (4.74E+00) - \\
    F6    & 5.77E-08 (1.34E-07) & 6.04E-09 (2.71E-08) = & 7.38E-09 (2.77E-08) = & 2.69E-09 (1.92E-08) = & 2.41E-08 (1.53E-07) = & 2.38E-13 (4.70E-14) - & 1.29E-11 (6.60E-11) - \\
\bottomrule
    \end{tabular}%
  \label{tab:cec17_30_wilcoxon}%
\footnotesize
\\The symbols ``+/=/-" indicate that the corresponding algorithm performed significantly better ($+$), not significantly better or worse ($=$), or significantly worse ($-$) compared to iLSHADE-RSP using the Wilcoxon rank-sum test with $\alpha = 0.05$ significance level.
\end{table*}%

\begin{table*}[h!]
  \tiny
  \centering
  \caption{Friedman test with Hochberg's post hoc for test algorithms on CEC 2017 test suite in 30 dimension}
    \begin{tabular}{cccccccccc}
    \toprule
          & Algorithm & Average ranking & z-value & p-value & Adj. p-value (Hochberg) & Sig.  &       & Test statistics &  \\
    \midrule
    1     & iLSHADE-RSP & 2.85  &       &       &       &       &       &       &  \\
    \cmidrule(l){2-3}
    2     & LSHADE-RSP & 3.47  & -6.62.E-01 & 5.08.E-01 & 1.02.E+00 & No    &       & N     & 30  \\
    3     & jSO   & 3.40  & -5.91.E-01 & 5.55.E-01 & 5.55.E-01 & No    &       & Chi-Square & 183.61  \\
    4     & L-SHADE & 3.62  & -8.24.E-01 & 4.10.E-01 & 1.23.E+00 & No    &       & df    & 11  \\
    5     & SHADE & 5.80  & -3.17.E+00 & 1.53.E-03 & 6.12.E-03 & Yes   &       & p-value & 1.84.E-33 \\
    6     & JADE  & 7.50  & -4.99.E+00 & 5.89.E-07 & 3.53.E-06 & Yes   &       & Sig.  & Yes \\
    7     & EDEV  & 7.53  & -5.03.E+00 & 4.89.E-07 & 3.42.E-06 & Yes   &       &       &  \\
    8     & MPEDE & 9.15  & -6.77.E+00 & 1.31.E-11 & 1.31.E-10 & Yes   &       &       &  \\
    9     & CoDE  & 10.90  & -8.65.E+00 & 5.28.E-18 & 5.81.E-17 & Yes   &       &       &  \\
    10    & EPSDE & 8.80  & -6.39.E+00 & 1.64.E-10 & 1.48.E-09 & Yes   &       &       &  \\
    11    & SaDE  & 8.70  & -6.28.E+00 & 3.30.E-10 & 2.64.E-09 & Yes   &       &       &  \\
    12    & dynNP-DE & 6.28  & -3.69.E+00 & 2.26.E-04 & 1.13.E-03 & Yes   &       &       &  \\
    \bottomrule
    \end{tabular}%
  \label{tab:cec17_30_friedman}%
\end{table*}%

\begin{table*}[h!]
  \tiny
  \centering
  \caption{Means and standard deviations of FEVs of test algorithms on CEC 2017 test suite in 50 dimension}
    \begin{tabular}{cccccccc}
    \toprule
          & iLSHADE-RSP & LSHADE-RSP & jSO   & L-SHADE & SHADE & JADE  & EDEV \\
          & MEAN (STD DEV) & MEAN (STD DEV) & MEAN (STD DEV) & MEAN (STD DEV) & MEAN (STD DEV) & MEAN (STD DEV) & MEAN (STD DEV) \\
    \midrule
    F1    & 2.03E-14 (1.21E-14) & 1.50E-14 (5.25E-15) = & 1.61E-14 (6.97E-15) = & 1.48E-14 (2.78E-15) = & 6.18E-14 (1.02E-13) - & 6.34E-14 (1.46E-13) - & 4.40E-12 (2.60E-11) = \\
    F2    & 9.75E-14 (3.30E-13) & 6.07E-14 (9.54E-14) = & 5.02E-14 (7.95E-14) = & 2.62E-14 (3.63E-14) + & 2.84E-09 (1.20E-08) - & 1.20E-09 (3.80E-09) - & 4.08E+27 (2.91E+28) - \\
    F3    & 2.13E-13 (7.51E-14) & 1.30E-13 (3.61E-14) + & 1.21E-13 (4.36E-14) + & 1.32E-13 (3.70E-14) + & 2.82E-13 (1.04E-13) - & 3.18E+04 (4.44E+04) - & 1.74E+04 (4.16E+04) + \\
    F4    & 4.76E+01 (4.52E+01) & 3.70E+01 (3.38E+01) - & 4.77E+01 (4.44E+01) - & 7.86E+01 (5.11E+01) - & 3.26E+01 (4.20E+01) = & 4.26E+01 (4.90E+01) = & 5.41E+01 (4.83E+01) = \\
    F5    & 1.68E+01 (5.19E+00) & 1.45E+01 (3.54E+00) + & 1.81E+01 (3.92E+00) = & 1.29E+01 (2.45E+00) + & 4.48E+01 (6.17E+00) - & 5.44E+01 (7.50E+00) - & 6.25E+01 (7.84E+00) - \\
    F6    & 7.10E-07 (1.04E-06) & 2.16E-07 (3.91E-07) + & 5.93E-07 (1.02E-06) = & 1.22E-04 (5.10E-04) + & 1.01E-05 (2.57E-05) - & 3.43E-13 (5.81E-14) + & 4.75E-13 (5.30E-13) + \\
\bottomrule
    \end{tabular}%
  \label{tab:cec17_50_wilcoxon}%
\footnotesize
\\The symbols ``+/=/-" indicate that the corresponding algorithm performed significantly better ($+$), not significantly better or worse ($=$), or significantly worse ($-$) compared to iLSHADE-RSP using the Wilcoxon rank-sum test with $\alpha = 0.05$ significance level.
\end{table*}%

\begin{table*}[h!]
  \tiny
  \centering
  \caption{Friedman test with Hochberg's post hoc for test algorithms on CEC 2017 test suite in 50 dimension}
    \begin{tabular}{cccccccccc}
    \toprule
          & Algorithm & Average ranking & z-value & p-value & Adj. p-value (Hochberg) & Sig.  &       & Test statistics &  \\
    \midrule
    1     & iLSHADE-RSP & 2.47  &       &       &       &       &       &       &  \\
    \cmidrule(l){2-3}
    2     & LSHADE-RSP & 2.63  & -1.79.E-01 & 8.58.E-01 & 8.58.E-01 & No    &       & N     & 30  \\
    3     & jSO   & 3.12  & -6.98.E-01 & 4.85.E-01 & 9.70.E-01 & No    &       & Chi-Square & 220.47  \\
    4     & L-SHADE & 3.73  & -1.36.E+00 & 1.74.E-01 & 5.21.E-01 & No    &       & df    & 11  \\
    5     & SHADE & 6.07  & -3.87.E+00 & 1.10.E-04 & 4.41.E-04 & Yes   &       & p-value & 4.12.E-41 \\
    6     & JADE  & 6.93  & -4.80.E+00 & 1.60.E-06 & 8.01.E-06 & Yes   &       & Sig.  & Yes \\
    7     & EDEV  & 7.43  & -5.34.E+00 & 9.55.E-08 & 6.69.E-07 & Yes   &       &       &  \\
    8     & MPEDE & 8.80  & -6.80.E+00 & 1.02.E-11 & 8.19.E-11 & Yes   &       &       &  \\
    9     & CoDE  & 11.40  & -9.60.E+00 & 8.32.E-22 & 9.15.E-21 & Yes   &       &       &  \\
    10    & EPSDE & 9.40  & -7.45.E+00 & 9.51.E-14 & 9.51.E-13 & Yes   &       &       &  \\
    11    & SaDE  & 8.93  & -6.95.E+00 & 3.75.E-12 & 3.37.E-11 & Yes   &       &       &  \\
    12    & dynNP-DE & 7.08  & -4.96.E+00 & 7.08.E-07 & 4.25.E-06 & Yes   &       &       &  \\
    \bottomrule
    \end{tabular}%
  \label{tab:cec17_50_friedman}%
\end{table*}%

\begin{table*}[h!]
  \tiny
  \centering
  \caption{Means and standard deviations of FEVs of test algorithms on CEC 2017 test suite in 100 dimension}
    \begin{tabular}{cccccccc}
    \toprule
          & iLSHADE-RSP & LSHADE-RSP & jSO   & L-SHADE & SHADE & JADE  & EDEV \\
          & MEAN (STD DEV) & MEAN (STD DEV) & MEAN (STD DEV) & MEAN (STD DEV) & MEAN (STD DEV) & MEAN (STD DEV) & MEAN (STD DEV) \\
    \midrule
    F1    & 9.60E-09 (1.39E-08) & 1.31E-10 (2.63E-10) + & 9.11E-11 (2.45E-10) + & 1.06E-12 (1.14E-12) + & 7.79E-10 (1.28E-09) + & 8.76E-10 (1.71E-09) + & 7.56E-10 (1.10E-09) + \\
    F2    & 2.75E+04 (1.50E+05) & 6.12E+06 (4.09E+07) = & 9.20E+04 (5.55E+05) + & 3.54E+11 (1.35E+12) - & 7.28E+24 (5.18E+25) = & 1.52E+22 (1.08E+23) = & 2.33E+26 (1.67E+27) + \\
    F3    & 8.60E-06 (7.87E-06) & 2.22E-06 (3.42E-06) + & 2.07E-06 (2.92E-06) + & 7.59E-07 (9.42E-07) + & 7.93E-02 (5.66E-01) + & 7.27E+04 (1.41E+05) + & 1.04E+05 (1.58E+05) + \\
    F4    & 2.00E+02 (2.43E+01) & 2.00E+02 (9.41E+00) + & 1.92E+02 (2.46E+01) + & 1.90E+02 (2.20E+01) + & 6.31E+01 (6.52E+01) + & 8.91E+01 (7.17E+01) + & 6.99E+01 (6.67E+01) + \\
    F5    & 3.73E+01 (1.26E+01) & 3.62E+01 (9.22E+00) = & 4.59E+01 (8.92E+00) - & 4.25E+01 (6.57E+00) - & 1.42E+02 (1.97E+01) - & 1.47E+02 (1.86E+01) - & 1.50E+02 (1.73E+01) - \\
    F6    & 3.68E-05 (2.07E-05) & 2.74E-05 (1.90E-05) + & 1.61E-04 (4.63E-04) = & 7.59E-03 (5.44E-03) - & 3.91E-02 (3.69E-02) - & 1.11E-03 (2.99E-03) + & 1.90E-04 (7.53E-04) + \\
\bottomrule
    \end{tabular}%
  \label{tab:cec17_100_wilcoxon}%
\footnotesize
\\The symbols ``+/=/-" indicate that the corresponding algorithm performed significantly better ($+$), not significantly better or worse ($=$), or significantly worse ($-$) compared to iLSHADE-RSP using the Wilcoxon rank-sum test with $\alpha = 0.05$ significance level.
\end{table*}%

\begin{table*}[h!]
  \tiny
  \centering
  \caption{Friedman test with Hochberg's post hoc for test algorithms on CEC 2017 test suite in 100 dimension}
    \begin{tabular}{cccccccccc}
    \toprule
          & Algorithm & Average ranking & z-value & p-value & Adj. p-value (Hochberg) & Sig.  &       & Test statistics &  \\
    \midrule
    1     & iLSHADE-RSP & 2.53  &       &       &       &       &       &       &  \\
    \cmidrule(l){2-3}
    2     & LSHADE-RSP & 2.67  & -1.43.E-01 & 8.86.E-01 & 8.86.E-01 & No    &       & N     & 30  \\
    3     & jSO   & 3.33  & -8.59.E-01 & 3.90.E-01 & 7.80.E-01 & No    &       & Chi-Square & 221.94  \\
    4     & L-SHADE & 3.77  & -1.32.E+00 & 1.85.E-01 & 5.56.E-01 & No    &       & df    & 11  \\
    5     & SHADE & 5.77  & -3.47.E+00 & 5.14.E-04 & 2.06.E-03 & Yes   &       & p-value & 2.04.E-41 \\
    6     & JADE  & 6.67  & -4.44.E+00 & 9.00.E-06 & 4.50.E-05 & Yes   &       & Sig.  & Yes \\
    7     & EDEV  & 7.33  & -5.16.E+00 & 2.52.E-07 & 1.51.E-06 & Yes   &       &       &  \\
    8     & MPEDE & 7.73  & -5.59.E+00 & 2.33.E-08 & 1.63.E-07 & Yes   &       &       &  \\
    9     & CoDE  & 11.87  & -1.00.E+01 & 1.18.E-23 & 1.29.E-22 & Yes   &       &       &  \\
    10    & EPSDE & 9.63  & -7.63.E+00 & 2.41.E-14 & 2.41.E-13 & Yes   &       &       &  \\
    11    & SaDE  & 8.33  & -6.23.E+00 & 4.66.E-10 & 3.73.E-09 & Yes   &       &       &  \\
    12    & dynNP-DE & 8.37  & -6.27.E+00 & 3.70.E-10 & 3.33.E-09 & Yes   &       &       &  \\
    \bottomrule
    \end{tabular}%
  \label{tab:cec17_100_friedman}%
\end{table*}%

\begin{figure*}[h!]
 \centering
 \subfigure[$F_{1}$]{
  \includegraphics[scale=0.33]{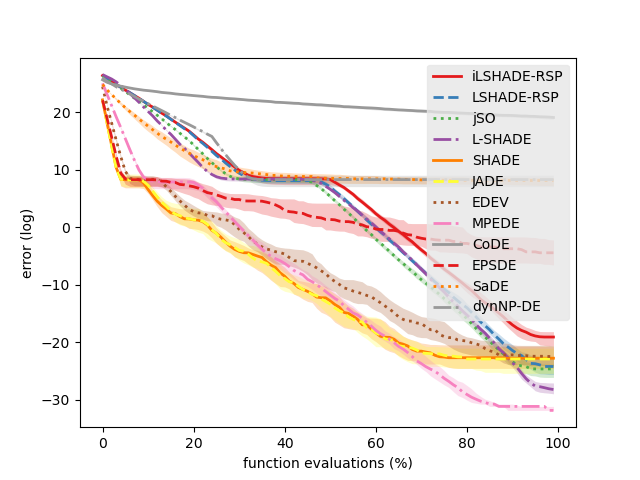}
   \label{fig:f1}
   }
 \subfigure[$F_{2}$]{
  \includegraphics[scale=0.33]{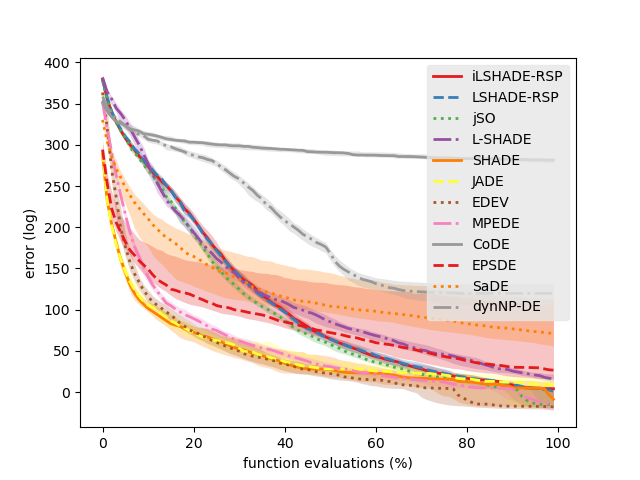}
   \label{fig:f2}
   }
 \subfigure[$F_{3}$]{
  \includegraphics[scale=0.33]{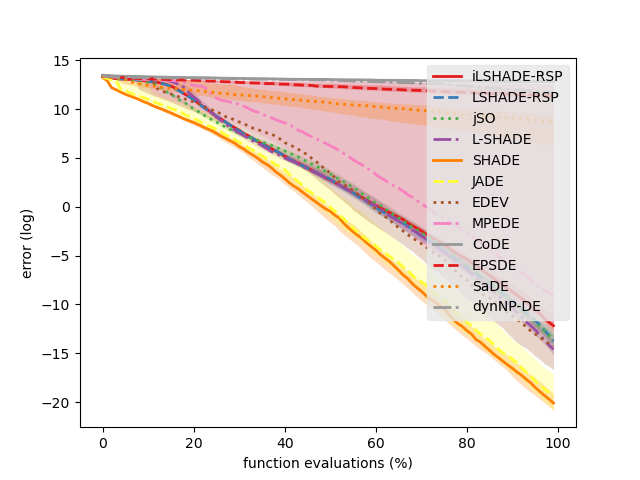}
   \label{fig:f3}
   }
 \subfigure[$F_{4}$]{
  \includegraphics[scale=0.33]{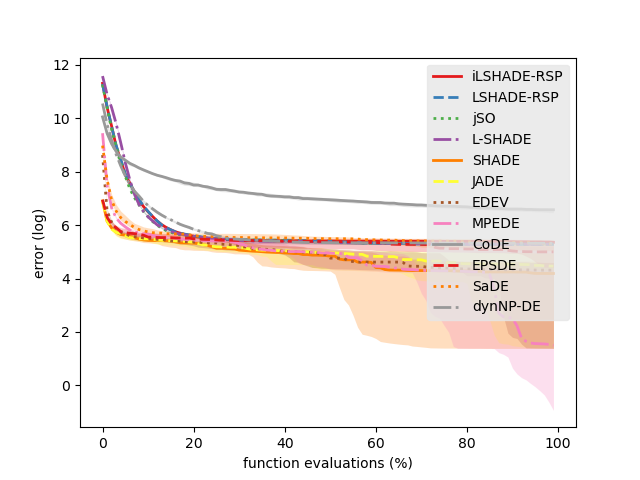}
   \label{fig:f4}
   }
 \subfigure[$F_{5}$]{
  \includegraphics[scale=0.33]{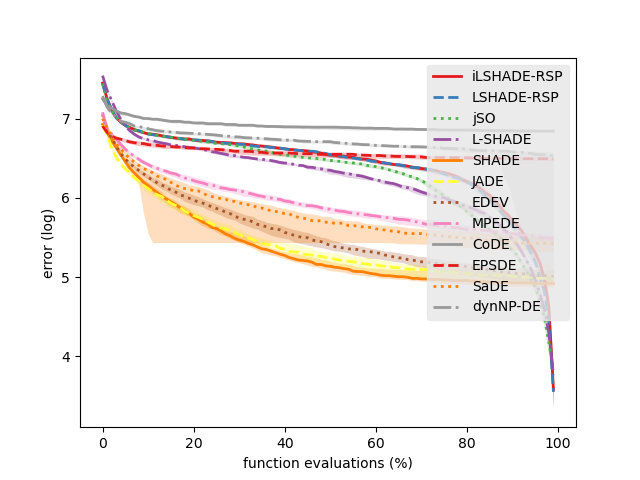}
   \label{fig:f5}
   }
 \subfigure[$F_{6}$]{
  \includegraphics[scale=0.33]{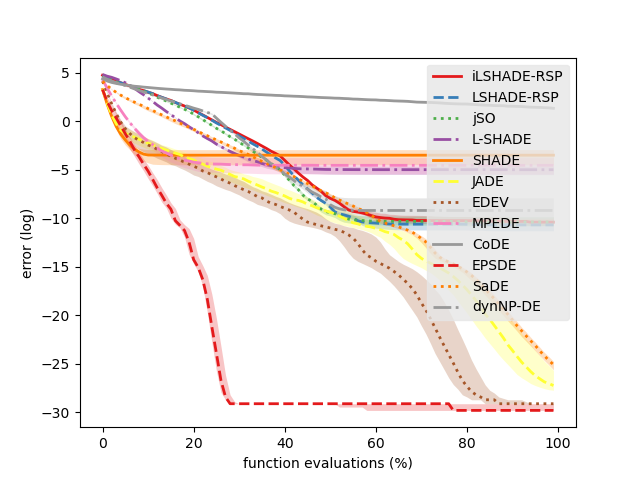}
   \label{fig:f6}
   }
 \caption[]{Convergence graphs of test algorithms on CEC 2017 test suites in 100 dimension ($F_{1}$ - $F_{6}$)}
 \label{fig:convergence_1}
\end{figure*}

\begin{figure*}[h!]
 \centering
 \subfigure[$F_{7}$]{
  \includegraphics[scale=0.33]{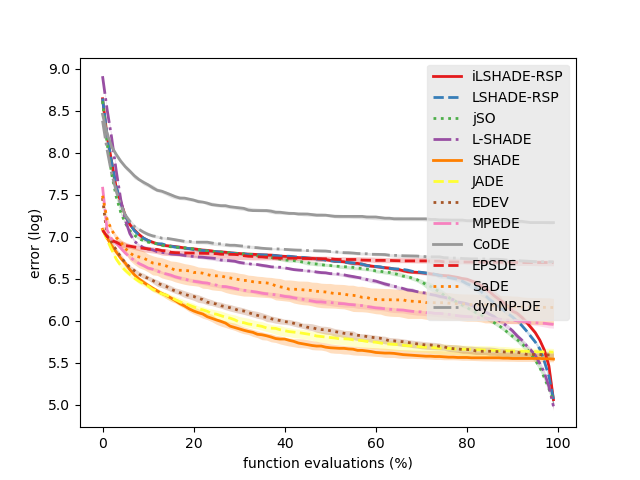}
   \label{fig:f7}
   }
 \subfigure[$F_{8}$]{
  \includegraphics[scale=0.33]{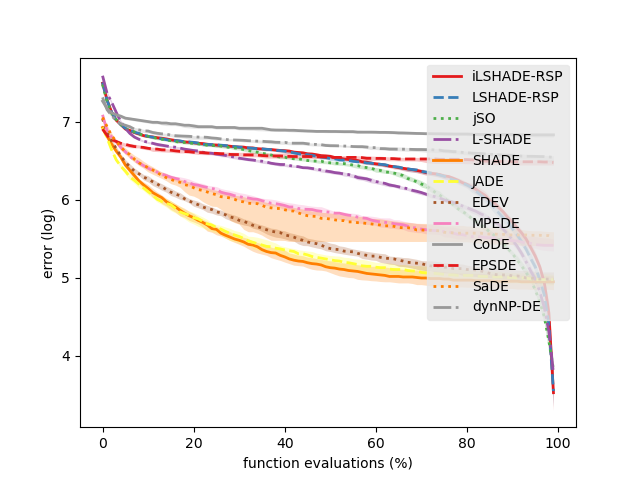}
   \label{fig:f8}
   }
 \subfigure[$F_{9}$]{
  \includegraphics[scale=0.33]{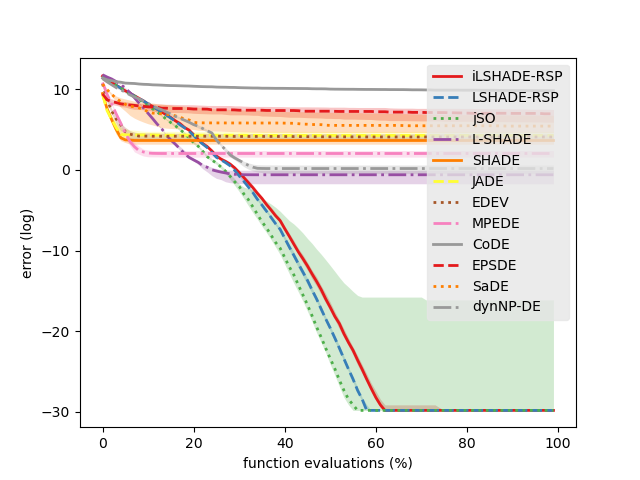}
   \label{fig:f9}
   }
 \subfigure[$F_{10}$]{
  \includegraphics[scale=0.33]{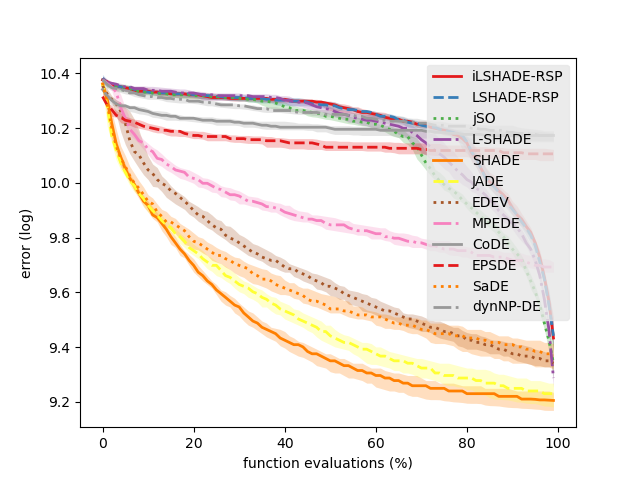}
   \label{fig:f10}
   }
 \subfigure[$F_{11}$]{
  \includegraphics[scale=0.33]{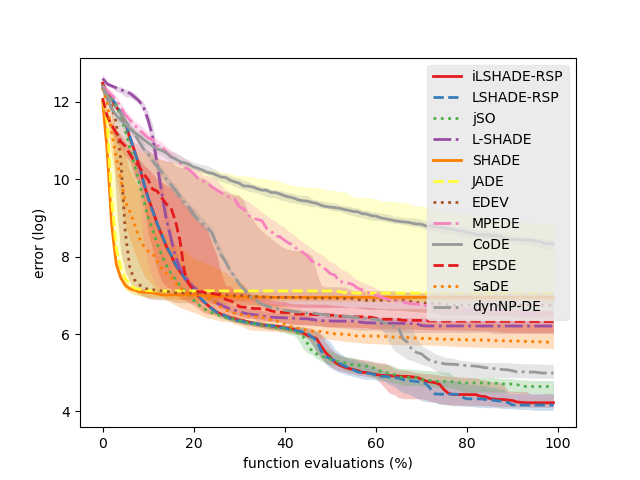}
   \label{fig:f11}
   }
 \subfigure[$F_{12}$]{
  \includegraphics[scale=0.33]{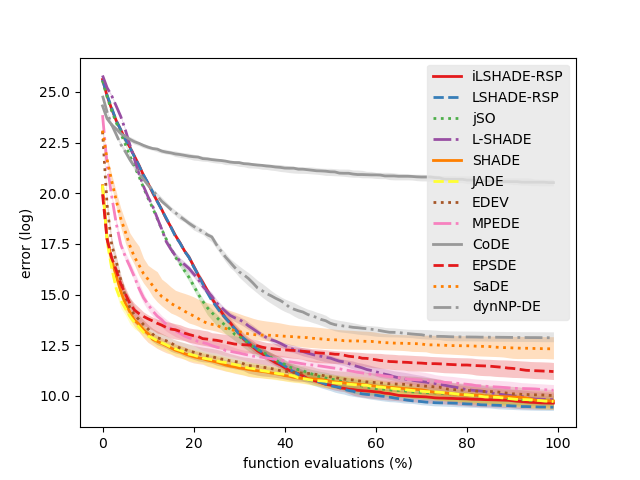}
   \label{fig:f12}
   }
 \caption[]{Convergence graphs of test algorithms on CEC 2017 test suites in 100 dimension ($F_{7}$ - $F_{12}$)}
 \label{fig:convergence_2}
\end{figure*}

\begin{figure*}[h!]
 \centering
 \subfigure[$F_{13}$]{
  \includegraphics[scale=0.33]{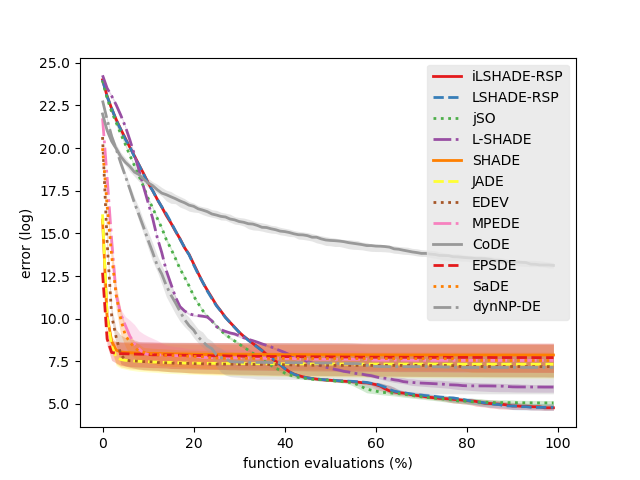}
   \label{fig:f13}
   }
 \subfigure[$F_{14}$]{
  \includegraphics[scale=0.33]{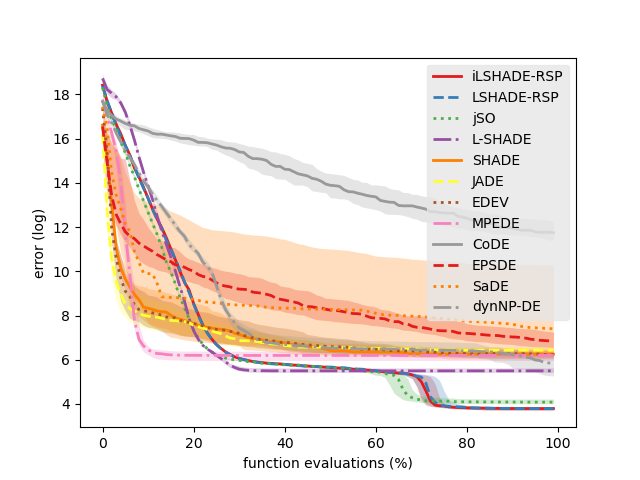}
   \label{fig:f14}
   }
 \subfigure[$F_{15}$]{
  \includegraphics[scale=0.33]{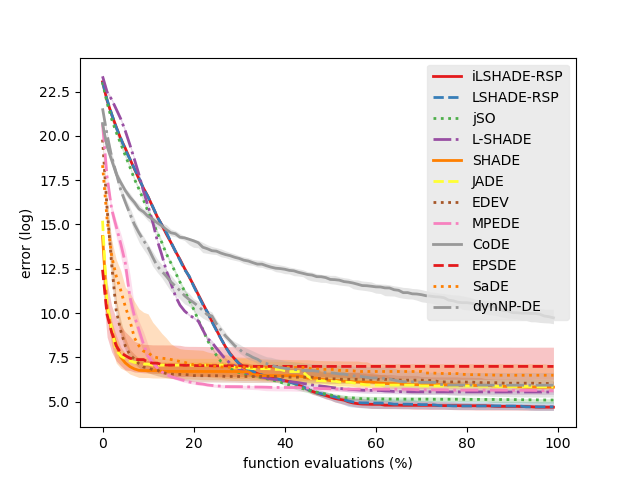}
   \label{fig:f15}
   }
 \subfigure[$F_{16}$]{
  \includegraphics[scale=0.33]{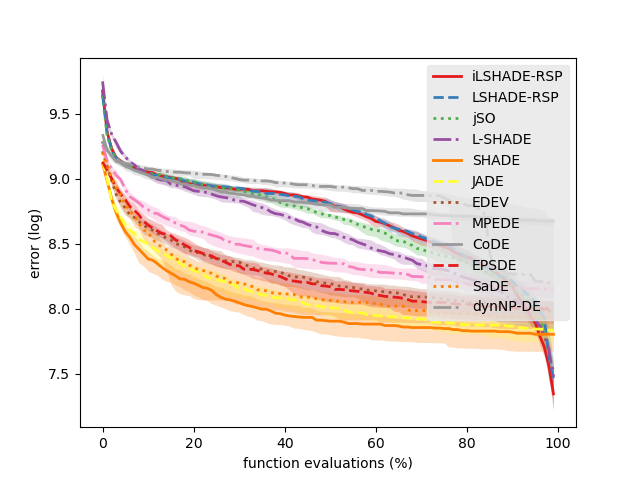}
   \label{fig:f16}
   }
 \subfigure[$F_{17}$]{
  \includegraphics[scale=0.33]{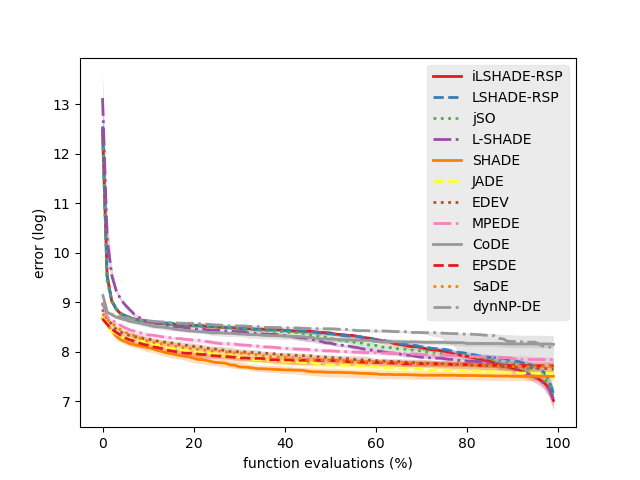}
   \label{fig:f17}
   }
 \subfigure[$F_{18}$]{
  \includegraphics[scale=0.33]{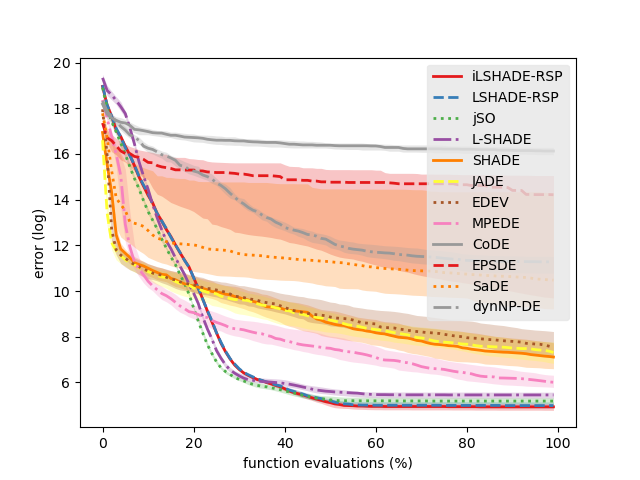}
   \label{fig:f18}
   }
 \caption[]{Convergence graphs of test algorithms on CEC 2017 test suites in 100 dimension ($F_{13}$ - $F_{18}$)}
 \label{fig:convergence_3}
\end{figure*}

\begin{figure*}[h!]
 \centering
 \subfigure[$F_{19}$]{
  \includegraphics[scale=0.33]{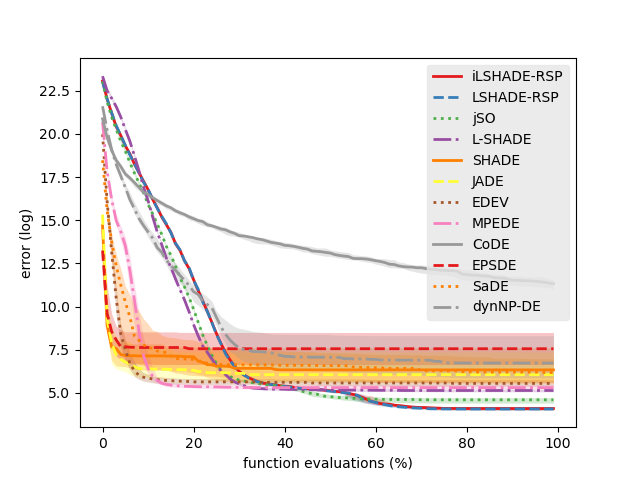}
   \label{fig:f19}
   }
 \subfigure[$F_{20}$]{
  \includegraphics[scale=0.33]{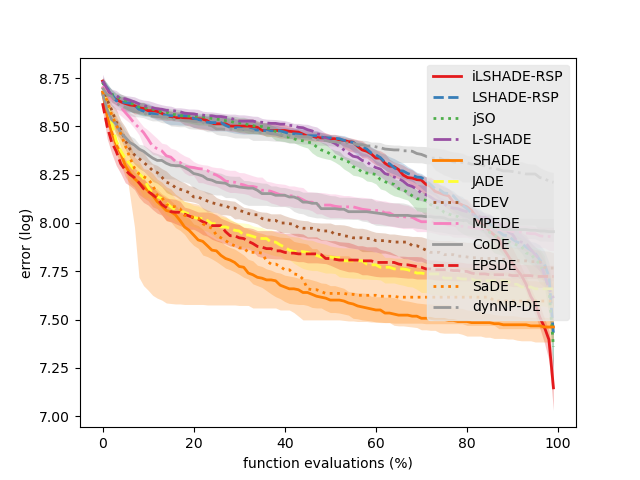}
   \label{fig:f20}
   }
 \subfigure[$F_{21}$]{
  \includegraphics[scale=0.33]{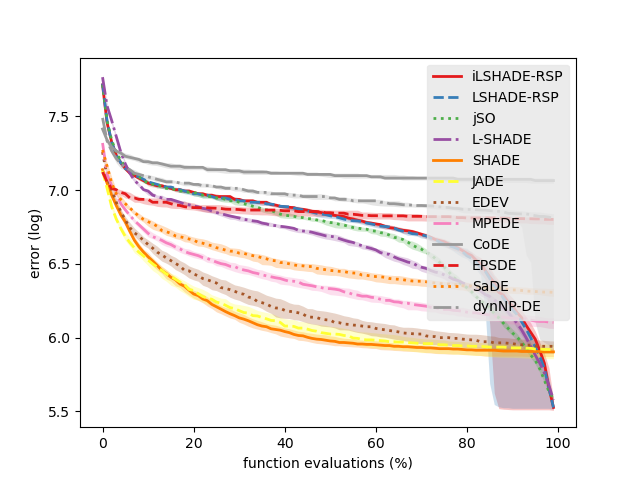}
   \label{fig:f21}
   }
 \subfigure[$F_{22}$]{
  \includegraphics[scale=0.33]{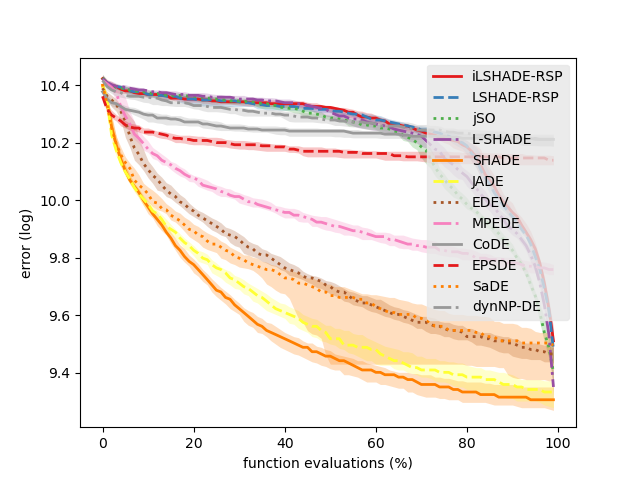}
   \label{fig:f22}
   }
 \subfigure[$F_{23}$]{
  \includegraphics[scale=0.33]{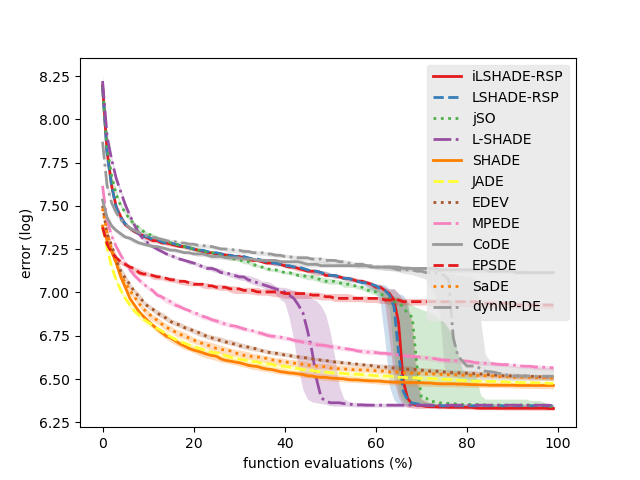}
   \label{fig:f23}
   }
 \subfigure[$F_{24}$]{
  \includegraphics[scale=0.33]{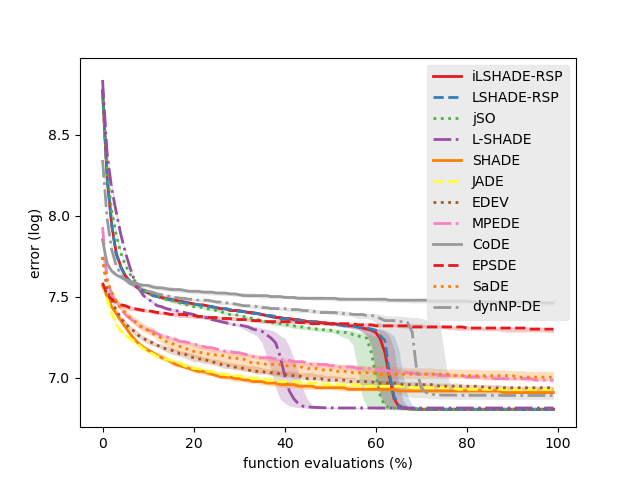}
   \label{fig:f24}
   }
 \caption[]{Convergence graphs of test algorithms on CEC 2017 test suites in 100 dimension ($F_{19}$ - $F_{24}$)}
 \label{fig:convergence_4}
\end{figure*}

\begin{figure*}[h!]
 \centering
 \subfigure[$F_{25}$]{
  \includegraphics[scale=0.33]{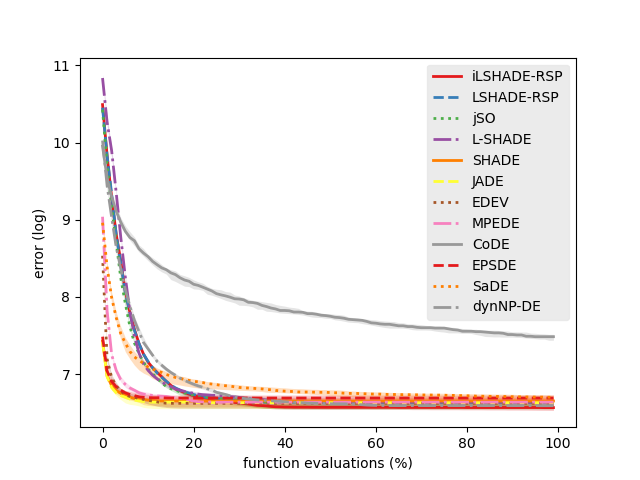}
   \label{fig:f25}
   }
 \subfigure[$F_{26}$]{
  \includegraphics[scale=0.33]{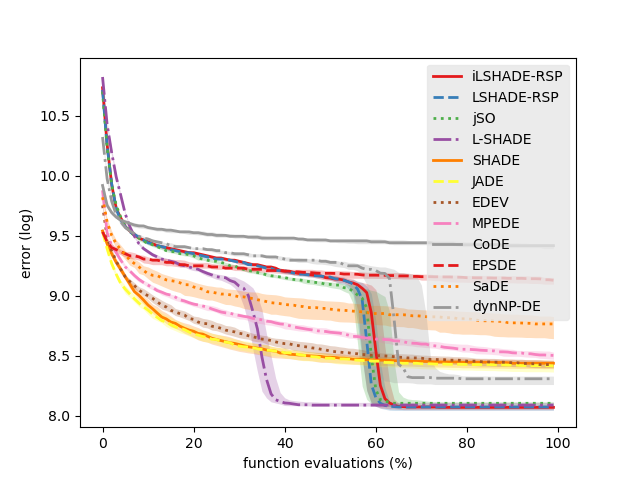}
   \label{fig:f26}
   }
 \subfigure[$F_{27}$]{
  \includegraphics[scale=0.33]{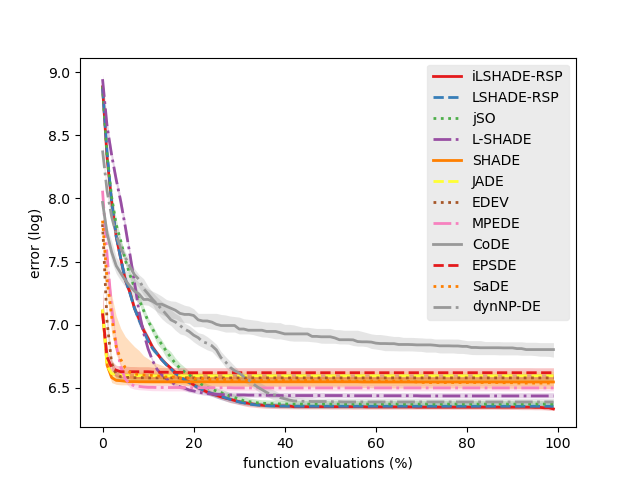}
   \label{fig:f27}
   }
 \subfigure[$F_{28}$]{
  \includegraphics[scale=0.33]{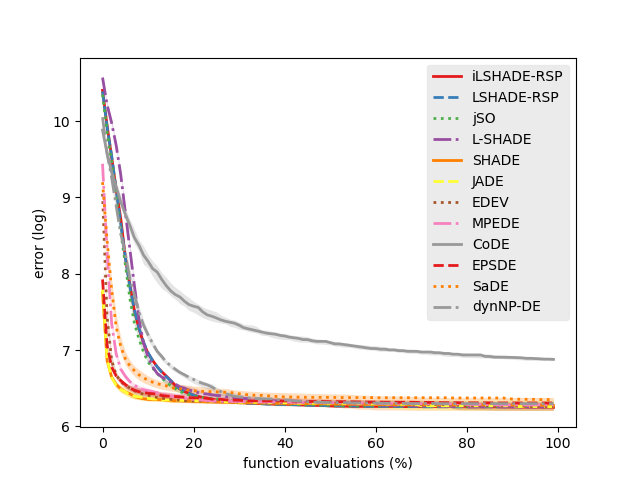}
   \label{fig:f28}
   }
 \subfigure[$F_{29}$]{
  \includegraphics[scale=0.33]{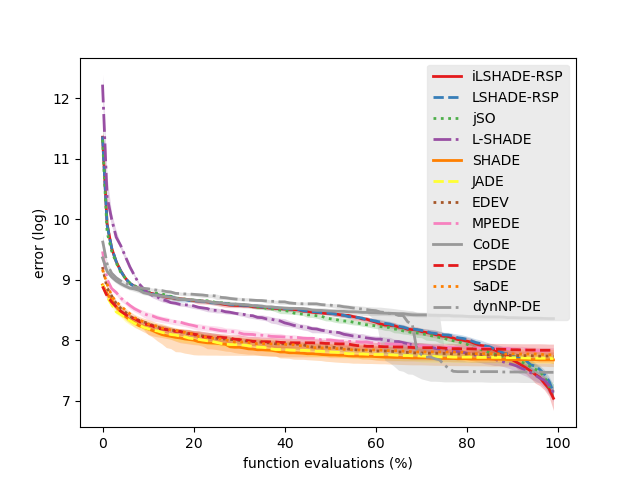}
   \label{fig:f29}
   }
 \subfigure[$F_{30}$]{
  \includegraphics[scale=0.33]{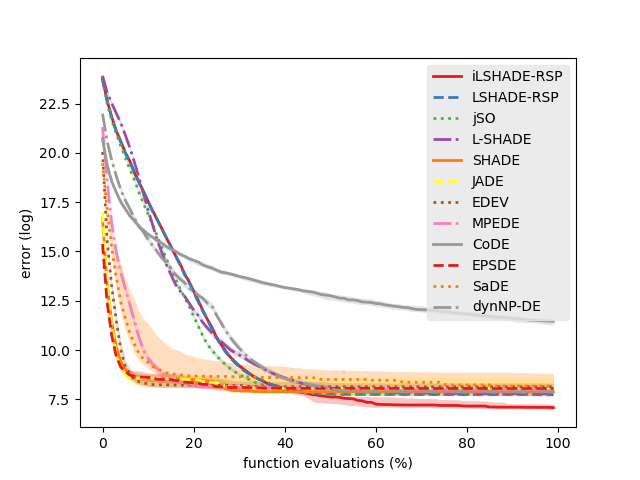}
   \label{fig:f30}
   }
 \caption[]{Convergence graphs of test algorithms on CEC 2017 test suites in 100 dimension ($F_{25}$ - $F_{30}$)}
 \label{fig:convergence_5}
\end{figure*}



In this section, we present the experimental results and discussion on the CEC 2017 test suite in 10, 30, 50, and 100 dimensions.

\subsection{Comparative Analysis}
We present the comparative analysis of the test algorithms in this subsection. Tables \ref{tab:cec17_10_wilcoxon}, \ref{tab:cec17_30_wilcoxon}, \ref{tab:cec17_50_wilcoxon}, and \ref{tab:cec17_100_wilcoxon} present the means and standard deviations of the FEVs of the test algorithms in 10, 30, 50, and 100 dimension, respectively. In the tables, the symbols ``+", ``=", and ``-" denote that the corresponding algorithm has statistically better, similar or worse performance compared to the proposed algorithm, respectively. Moreover, Tables \ref{tab:cec17_10_friedman}, \ref{tab:cec17_30_friedman}, \ref{tab:cec17_50_friedman}, and \ref{tab:cec17_100_friedman} present the results of the Friedman test with Hochberg's post hoc in 10, 30, 50, and 100 dimension, respectively. Furthermore, Figs. \ref{fig:convergence_1}, \ref{fig:convergence_2}, \ref{fig:convergence_3}, \ref{fig:convergence_4}, and \ref{fig:convergence_5} provide the convergence graphs of the test algorithms in 100 dimension.

Table \ref{tab:cec17_10_wilcoxon} presents the means and standard deviations of the FEVs of the test algorithms in 10 dimension, obtained by 51 independent runs. As can be seen from the table, the proposed algorithm performs better performance than all of the other test algorithms. Specifically, iLSHADE-RSP found significantly better solutions with lower FEVs than SHADE, JADE, EDEV, MPEDE, CoDE, EPSDE, and SaDE on more than 50 percent of the test functions. In particular, MPEDE and CoDE were significantly outperformed by iLSHADE-RSP on approximately 80 percent of the test functions. As compared to its predecessor LSHADE-RSP, the proposed algorithm considerably outperformed on 2 test functions and underperformed it on 0 test functions. In addition, Table \ref{tab:cec17_10_friedman} presents the Friedman test with Hochberg's post hoc, which supports the comparative analysis in Table \ref{tab:cec17_10_wilcoxon} where iLSHADE-RSP ranked the first among the test algorithms, and the outperformance over JADE, EDEV, MPEDE, CoDE, EPSDE, and SaDE was statistically significant.

The means and standard deviations of the FEVs of the proposed and comparison algorithms in 30 dimension are shown in Table \ref{tab:cec17_30_wilcoxon}, collected by 51 independent runs. As can be seen from the table, the proposed algorithm performs better performance than all of the other test algorithms. Specifically, iLSHADE-RSP found significantly better solutions with lower FEVs than SHADE, JADE, EDEV, MPEDE, CoDE, EPSDE, SaDE, and dynNP-DE on more than 80 percent of the test functions. In particular, JADE, EDEV, MPEDE, CoDE, EPSDE, and SaDE were not able to outperform iLSHADE-RSP on any of the test functions. As compared to its predecessor LSHADE-RSP, the proposed algorithm considerably outperformed on 6 test functions and underperformed it on 1 test functions. Additionally, Table \ref{tab:cec17_30_friedman} presents the Friedman test with Hochberg's post hoc, which supports the comparative analysis in Table \ref{tab:cec17_30_wilcoxon} where iLSHADE-RSP ranked the first among the test algorithms, and the outperformance over SHADE, JADE, EDEV, MPEDE, CoDE, EPSDE, SaDE, and dynNP-DE was statistically significant.

Table \ref{tab:cec17_50_wilcoxon} presents the means and standard deviations of the FEVs of the test algorithms in 50 dimension, obtained by 51 independent runs. As can be seen from the table, the proposed algorithm performs better performance than all of the other test algorithms. Specifically, iLSHADE-RSP found significantly better solutions with lower FEVs than all of the other test algorithms except LSHADE-RSP on more than 50 percent of the test functions. In particular, SHADE, JADE, EDEV, MPEDE, CoDE, EPSDE, SaDE, and dynNP-DE were significantly outperformed by iLSHADE-RSP on approximately 90 percent of the test functions. As compared to its predecessor LSHADE-RSP, the proposed algorithm considerably outperformed on 8 test functions and underperformed it on 4 test functions. In addition, Table \ref{tab:cec17_50_friedman} presents the Friedman test with Hochberg's post hoc, which supports the comparative analysis in Table \ref{tab:cec17_50_wilcoxon} where iLSHADE-RSP ranked the first among the test algorithms, and the outperformance over SHADE, JADE, EDEV, MPEDE, CoDE, EPSDE, SaDE, and dynNP-DE was statistically significant.

The means and standard deviations of the FEVs of the proposed and comparison algorithms in 100 dimension are shown in Table \ref{tab:cec17_100_wilcoxon}, collected by 51 independent runs. As can be seen from the table, the proposed algorithm performs better performance than all of the other test algorithms. Specifically, iLSHADE-RSP found significantly better solutions with lower FEVs than all of the other test algorithms except LSHADE-RSP on more than 50 percent of the test functions. In particular, MPEDE, CoDE, EPSDE, SaDE, and dynNP-DE were significantly outperformed by iLSHADE-RSP on approximately 90 percent of the test functions. As compared to its predecessor LSHADE-RSP, the proposed algorithm considerably outperformed on 8 test functions and underperformed it on 5 test functions. Additionally, Table \ref{tab:cec17_100_friedman} presents the Friedman test with Hochberg's post hoc, which supports the comparative analysis in Table \ref{tab:cec17_100_wilcoxon} where iLSHADE-RSP ranked the first among the test algorithms, and the outperformance over SHADE, JADE, EDEV, MPEDE, CoDE, EPSDE, SaDE, and dynNP-DE was statistically significant.

The convergence graphs of the proposed and comparison algorithms in 100 dimension are provided in Figs \ref{fig:convergence_1}, \ref{fig:convergence_2}, \ref{fig:convergence_3}, \ref{fig:convergence_4}, and \ref{fig:convergence_5}. Each convergence graph provides the median and interquartile ranges (one-fourth and three-fourth) of the FEVs of the proposed and comparison algorithms. The first part of the convergence graphs ($F_{1}$ - $F_{6}$) is given in Fig. \ref{fig:convergence_1}. The second part of the convergence graphs ($F_{7}$ - $F_{12}$) is given in Fig. \ref{fig:convergence_2}. The third part of the convergence graphs ($F_{13}$ - $F_{18}$) is given in Fig. \ref{fig:convergence_3}. The fourth part of the convergence graphs ($F_{19}$ - $F_{24}$) is given in Fig. \ref{fig:convergence_4}. The last part of the convergence graphs ($F_{25}$ - $F_{30}$) is given in Fig. \ref{fig:convergence_5}. As can be seen from the figures, iLSHADE-RSP is competitive with the other test algorithms in terms of solution accuracy, especially for the test functions $F_{5}$, $F_{7}$ - $F_{9}$, $F_{11}$ - $F_{21}$, and $F_{23}$ - $F_{30}$. In particular, iLSHADE-RSP can escape from the local optimum of the test functions $F_{20}$ and $F_{30}$, while the other test algorithms cannot.

\subsection{Discussion of Comparative Analysis}
We first discuss the comparative analysis between the proposed algorithm and each of the two state-of-the-art L-SHADE variants, LSHADE-RSP and jSO. After that, we discuss the experimental results with the proposed algorithm and the other test algorithms.

\begin{itemize}
    \item The performance difference between the proposed algorithm and LSHADE-RSP is negligible in 10 dimension. The proposed algorithm has only two improvements against zero deteriorations. However, the performance difference is much different in 30, 50, and 100 dimensions. The proposed algorithm has six improvements against one deterioration in 30 dimension, eight improvements against four deteriorations in 50 dimension, and eight improvements against five deteriorations in 100 dimension. Additionally, we investigated the performance difference with respect to the characteristics of the test functions. We found out that the proposed algorithm has worse performance on the unimodal ($F_{1}$-$F_{3}$) and some of the simple multimodal test functions ($F_{4}$-$F_{10}$) but better performance on the expanded multimodal ($F_{11}$-$F_{20}$) and the hybrid composition test functions ($F_{21}$-$F_{30}$). In a word, the proposed algorithm performs better than LSHADE-RSP on more complicated optimization problems.
    \item The performance difference between the proposed algorithm and jSO is negligible in 10 dimension. The proposed algorithm has only five improvements against three deteriorations. However, the performance difference is much different in 30, 50, and 100 dimensions. The proposed algorithm has 11 improvements against six deteriorations in 30 dimension, 17 improvements against two deteriorations in 50 dimension, and 18 improvements against six deteriorations in 100 dimension. Additionally, we investigated the performance difference with respect to the characteristics of the test functions. We found out that the proposed algorithm has worse performance on the unimodal ($F_{1}$-$F_{3}$) and some of the simple multimodal test functions ($F_{4}$-$F_{10}$) but better performance on the expanded multimodal ($F_{11}$-$F_{20}$) and the hybrid composition test functions ($F_{21}$-$F_{30}$). In a word, the proposed algorithm performs better than jSO on more complicated optimization problems.
    \item The proposed algorithm found more significantly accurate solutions compared with the test algorithms, including L-SHADE, SHADE, JADE, EDEV, MPEDE, CoDE, EPSDE, SaDE, and dynNP-DE, in all the dimensions.
\end{itemize}

Note that, the experimental results with the proposed algorithm and LSHADE-RSP lend weight to the effectiveness of the modified recombination operator of the proposed algorithm, which can increase the probability of finding an optimal solution by adopting the long-tailed property of the Cauchy distribution, and thus, it can improve the optimization performance of LSHADE-RSP significantly.


\begin{table*}[h!]
  \tiny
  \centering
  \caption{Means and standard deviations of FEVs of iLSHADE-RSP with different jumping rates on CEC 2017 test suite in 50 dimension}
    \begin{tabular}{ccccccccc}
    \toprule
          & iLSHADE-RSP &       &       &       &       &       &  \\
    \cmidrule(l){2-2}
          & $p_{j}=0.20$ & $p_{j}=0.05$ & $p_{j}=0.10$ & $p_{j}=0.15$ & $p_{j}=0.25$ & $p_{j}=0.30$ & $p_{j}=0.35$ & $p_{j}=0.40$ \\
          & MEAN (STD DEV) & MEAN (STD DEV) & MEAN (STD DEV) & MEAN (STD DEV) & MEAN (STD DEV) & MEAN (STD DEV) & MEAN (STD DEV) & MEAN (STD DEV) \\
    \midrule
    F1    & 2.26E-14 (1.99E-14) & 1.56E-14 (6.51E-15) + & 1.61E-14 (6.36E-15) = & 1.98E-14 (2.12E-14) = & 2.65E-14 (3.32E-14) = & 2.92E-14 (2.82E-14) = & 3.90E-14 (4.52E-14) - & 6.05E-14 (9.22E-14) - \\
    F2    & 8.13E-14 (2.21E-13) & 4.29E-14 (5.25E-14) = & 3.85E-13 (2.17E-12) = & 6.35E-14 (1.48E-13) = & 7.47E-14 (1.33E-13) = & 1.55E-13 (6.30E-13) = & 5.52E-14 (7.29E-14) = & 1.27E-13 (2.28E-13) = \\
    F3    & 2.01E-13 (8.60E-14) & 1.39E-13 (4.14E-14) + & 1.66E-13 (5.98E-14) + & 1.73E-13 (5.43E-14) = & 2.47E-13 (9.49E-14) - & 2.91E-13 (1.24E-13) - & 4.28E-13 (1.84E-13) - & 4.87E-13 (2.28E-13) - \\
    F4    & 6.83E+01 (5.32E+01) & 5.61E+01 (4.73E+01) = & 5.45E+01 (5.10E+01) = & 6.55E+01 (5.32E+01) = & 5.62E+01 (5.03E+01) + & 6.04E+01 (5.20E+01) + & 7.08E+01 (5.44E+01) + & 5.54E+01 (4.96E+01) + \\
    F5    & 1.51E+01 (5.06E+00) & 1.50E+01 (3.86E+00) = & 1.49E+01 (3.78E+00) = & 1.52E+01 (4.47E+00) = & 1.56E+01 (4.96E+00) = & 1.60E+01 (5.66E+00) = & 1.53E+01 (5.10E+00) = & 1.65E+01 (5.35E+00) = \\
    F6    & 9.98E-07 (1.80E-06) & 2.61E-07 (5.54E-07) + & 3.93E-07 (5.74E-07) = & 5.26E-07 (9.13E-07) = & 7.25E-07 (1.27E-06) = & 1.22E-06 (1.53E-06) - & 2.11E-06 (3.76E-06) - & 3.62E-06 (3.35E-06) - \\
    F7    & 7.25E+01 (7.22E+00) & 6.96E+01 (5.97E+00) + & 7.15E+01 (5.70E+00) = & 7.11E+01 (6.27E+00) = & 7.37E+01 (8.33E+00) = & 7.55E+01 (7.89E+00) = & 7.63E+01 (9.80E+00) = & 7.60E+01 (1.00E+01) = \\
    F8    & 1.62E+01 (4.46E+00) & 1.64E+01 (5.05E+00) = & 1.55E+01 (3.86E+00) = & 1.56E+01 (5.44E+00) = & 1.74E+01 (5.45E+00) = & 1.74E+01 (5.56E+00) = & 1.67E+01 (5.28E+00) = & 1.66E+01 (4.91E+00) = \\
    F9    & 4.25E-14 (5.57E-14) & 6.71E-15 (2.71E-14) + & 2.24E-14 (4.57E-14) = & 2.24E-14 (4.57E-14) = & 3.58E-14 (5.34E-14) = & 4.69E-14 (5.67E-14) = & 4.47E-14 (5.62E-14) = & 5.81E-14 (5.76E-14) = \\
    F10   & 4.23E+03 (5.00E+02) & 4.13E+03 (5.09E+02) = & 4.18E+03 (6.67E+02) = & 4.24E+03 (6.18E+02) = & 4.25E+03 (6.39E+02) = & 4.19E+03 (6.33E+02) = & 4.20E+03 (5.26E+02) = & 4.13E+03 (5.96E+02) = \\
    F11   & 1.77E+01 (3.42E+00) & 1.97E+01 (4.26E+00) - & 1.75E+01 (3.65E+00) = & 1.74E+01 (4.07E+00) = & 1.82E+01 (3.28E+00) = & 1.87E+01 (2.67E+00) = & 1.82E+01 (2.74E+00) = & 1.89E+01 (2.76E+00) - \\
    F12   & 1.42E+03 (4.59E+02) & 1.55E+03 (4.16E+02) = & 1.45E+03 (4.04E+02) = & 1.43E+03 (4.25E+02) = & 1.43E+03 (4.15E+02) = & 1.47E+03 (4.64E+02) = & 1.39E+03 (3.50E+02) = & 1.41E+03 (3.96E+02) = \\
    F13   & 3.03E+01 (2.09E+01) & 2.38E+01 (1.68E+01) = & 2.91E+01 (2.44E+01) = & 2.90E+01 (2.34E+01) = & 2.99E+01 (2.42E+01) = & 2.83E+01 (1.93E+01) = & 3.09E+01 (1.73E+01) = & 3.03E+01 (1.91E+01) = \\
    F14   & 2.36E+01 (1.88E+00) & 2.40E+01 (2.17E+00) = & 2.40E+01 (2.02E+00) = & 2.33E+01 (2.07E+00) = & 2.36E+01 (2.09E+00) = & 2.34E+01 (2.00E+00) = & 2.36E+01 (1.82E+00) = & 2.34E+01 (1.76E+00) = \\
    F15   & 1.87E+01 (1.89E+00) & 1.91E+01 (1.86E+00) = & 1.97E+01 (1.61E+00) - & 1.88E+01 (1.63E+00) = & 1.88E+01 (2.48E+00) = & 1.83E+01 (1.72E+00) = & 1.84E+01 (2.14E+00) = & 1.84E+01 (1.71E+00) = \\
    F16   & 2.94E+02 (1.29E+02) & 3.04E+02 (1.39E+02) = & 3.15E+02 (1.28E+02) = & 3.19E+02 (1.37E+02) = & 3.03E+02 (1.12E+02) = & 3.20E+02 (1.18E+02) = & 3.48E+02 (1.32E+02) - & 3.35E+02 (1.16E+02) - \\
    F17   & 2.27E+02 (8.62E+01) & 2.34E+02 (9.75E+01) = & 2.42E+02 (1.06E+02) = & 2.47E+02 (9.24E+01) = & 2.29E+02 (8.94E+01) = & 2.30E+02 (8.92E+01) = & 2.36E+02 (1.06E+02) = & 2.47E+02 (9.12E+01) = \\
    F18   & 2.26E+01 (1.27E+00) & 2.27E+01 (1.27E+00) = & 2.29E+01 (1.41E+00) = & 2.27E+01 (1.18E+00) = & 2.30E+01 (1.46E+00) = & 2.27E+01 (1.49E+00) = & 2.25E+01 (1.32E+00) = & 2.26E+01 (1.40E+00) = \\
    F19   & 1.03E+01 (2.46E+00) & 1.13E+01 (2.01E+00) - & 1.06E+01 (2.31E+00) = & 1.04E+01 (2.85E+00) = & 1.06E+01 (2.12E+00) = & 1.10E+01 (2.26E+00) = & 1.07E+01 (2.43E+00) = & 1.10E+01 (2.35E+00) = \\
    F20   & 1.18E+02 (4.79E+01) & 1.40E+02 (7.51E+01) = & 1.18E+02 (5.10E+01) = & 1.17E+02 (2.78E+01) = & 1.17E+02 (4.85E+01) = & 1.27E+02 (4.78E+01) = & 1.23E+02 (4.14E+01) = & 1.35E+02 (5.64E+01) - \\
    F21   & 2.15E+02 (4.35E+00) & 2.15E+02 (4.13E+00) = & 2.15E+02 (3.70E+00) = & 2.16E+02 (5.24E+00) = & 2.16E+02 (4.93E+00) = & 2.17E+02 (5.49E+00) = & 2.15E+02 (5.17E+00) = & 2.17E+02 (4.96E+00) = \\
    F22   & 1.86E+03 (2.25E+03) & 1.91E+03 (2.21E+03) = & 1.58E+03 (2.22E+03) = & 1.51E+03 (2.13E+03) = & 1.86E+03 (2.25E+03) = & 1.31E+03 (2.01E+03) = & 1.70E+03 (2.13E+03) = & 1.62E+03 (2.14E+03) = \\
    F23   & 4.33E+02 (7.17E+00) & 4.31E+02 (5.46E+00) = & 4.33E+02 (6.28E+00) = & 4.33E+02 (7.21E+00) = & 4.32E+02 (6.43E+00) = & 4.32E+02 (6.17E+00) = & 4.33E+02 (6.92E+00) = & 4.33E+02 (6.58E+00) = \\
    F24   & 5.09E+02 (4.10E+00) & 5.08E+02 (3.48E+00) = & 5.09E+02 (4.42E+00) = & 5.09E+02 (3.67E+00) = & 5.10E+02 (4.09E+00) = & 5.10E+02 (4.15E+00) = & 5.10E+02 (4.31E+00) = & 5.09E+02 (3.48E+00) = \\
    F25   & 4.79E+02 (9.87E-01) & 4.80E+02 (1.41E+00) - & 4.80E+02 (7.77E-01) - & 4.80E+02 (8.57E-01) = & 4.79E+02 (8.08E-01) = & 4.79E+02 (9.61E-01) = & 4.79E+02 (9.51E-01) = & 4.79E+02 (7.84E-01) + \\
    F26   & 1.13E+03 (4.73E+01) & 1.15E+03 (5.73E+01) = & 1.13E+03 (5.44E+01) = & 1.14E+03 (5.75E+01) = & 1.13E+03 (5.27E+01) = & 1.12E+03 (5.25E+01) + & 1.13E+03 (5.84E+01) = & 1.13E+03 (5.18E+01) = \\
    F27   & 4.77E+02 (6.46E+00) & 4.89E+02 (9.29E+00) - & 4.83E+02 (9.11E+00) - & 4.78E+02 (5.88E+00) = & 4.79E+02 (7.93E+00) = & 4.75E+02 (5.64E+00) = & 4.74E+02 (5.90E+00) + & 4.74E+02 (8.15E+00) + \\
    F28   & 4.52E+02 (3.37E-01) & 4.53E+02 (6.69E-01) - & 4.53E+02 (5.94E-01) - & 4.52E+02 (3.82E-01) = & 4.52E+02 (4.81E+00) = & 4.52E+02 (4.76E-01) + & 4.52E+02 (4.76E-01) + & 4.51E+02 (4.93E-01) + \\
    F29   & 3.06E+02 (1.93E+01) & 3.28E+02 (3.69E+01) - & 3.17E+02 (2.52E+01) - & 3.12E+02 (2.03E+01) = & 3.04E+02 (2.04E+01) = & 2.99E+02 (1.75E+01) + & 3.01E+02 (1.84E+01) = & 2.99E+02 (1.61E+01) + \\
    F30   & 4.19E+03 (5.95E+03) & 1.76E+04 (4.90E+04) - & 3.97E+03 (6.15E+03) = & 2.79E+03 (3.05E+03) = & 2.68E+03 (2.65E+03) = & 3.79E+03 (3.31E+03) = & 6.32E+03 (7.45E+03) = & 6.26E+03 (6.91E+03) - \\
    \midrule
    +/=/- &       & 5/18/7 & 1/24/5 & 0/30/0 & 1/28/1 & 4/24/2 & 3/23/4 & 5/18/7 \\
    \bottomrule
    \end{tabular}%
  \label{tab:JumpingRate}%
\footnotesize
\\The symbols ``+/=/-" indicate that the corresponding algorithm performed significantly better ($+$), not significantly better or worse ($=$), or significantly worse ($-$) compared to iLSHADE-RSP with $p_{j} = 0.2$ using the Wilcoxon rank-sum test with $\alpha = 0.05$ significance level.
\end{table*}%

\begin{table*}[h!]
  \tiny
  \centering
  \caption{Means and standard deviations of FEVs of iLSHADE-RSP with different stability parameters on CEC 2017 test suite in 50 dimension}
    \begin{tabular}{ccccccccc}
    \toprule
          & iLSHADE-RSP &       &       &       &       &       &  \\
    \cmidrule(l){2-2}
          & $\alpha=1.0$ & $\alpha=0.3$ & $\alpha=0.5$ & $\alpha=0.8$ & $\alpha=1.3$ & $\alpha=1.5$ & $\alpha=1.8$ & $\alpha=2.0$ \\
          & MEAN (STD DEV) & MEAN (STD DEV) & MEAN (STD DEV) & MEAN (STD DEV) & MEAN (STD DEV) & MEAN (STD DEV) & MEAN (STD DEV) & MEAN (STD DEV) \\
    \midrule
    F1    & 2.03E-14 (1.43E-14) & 2.09E-14 (1.25E-14) = & 2.09E-14 (1.37E-14) = & 2.03E-14 (8.64E-15) = & 1.73E-14 (6.55E-15) = & 2.12E-14 (1.28E-14) = & 2.17E-14 (1.62E-14) = & 1.92E-14 (1.23E-14) = \\
    F2    & 6.18E-14 (1.07E-13) & 4.21E-11 (3.00E-10) = & 1.29E-13 (6.31E-13) = & 1.67E-13 (7.37E-13) = & 1.25E-13 (3.92E-13) = & 4.51E-14 (9.51E-14) = & 1.59E-13 (5.81E-13) = & 4.46E-14 (4.84E-14) = \\
    F3    & 2.29E-13 (6.66E-14) & 2.17E-13 (7.92E-14) = & 1.95E-13 (6.63E-14) + & 2.15E-13 (1.00E-13) = & 1.86E-13 (6.32E-14) + & 2.01E-13 (5.92E-14) + & 2.20E-13 (6.41E-14) = & 1.94E-13 (6.01E-14) + \\
    F4    & 5.93E+01 (5.19E+01) & 7.70E+01 (5.40E+01) - & 6.12E+01 (5.24E+01) - & 6.44E+01 (5.28E+01) - & 5.09E+01 (4.67E+01) + & 6.34E+01 (5.08E+01) = & 7.00E+01 (5.30E+01) = & 5.28E+01 (4.87E+01) = \\
    F5    & 1.50E+01 (5.08E+00) & 1.55E+01 (4.26E+00) = & 1.57E+01 (5.91E+00) = & 1.69E+01 (3.98E+00) - & 1.48E+01 (4.27E+00) = & 1.46E+01 (4.74E+00) = & 1.36E+01 (3.66E+00) = & 1.52E+01 (4.07E+00) = \\
    F6    & 4.24E-07 (5.09E-07) & 4.60E-07 (6.69E-07) = & 5.48E-07 (8.21E-07) = & 7.05E-07 (1.02E-06) = & 5.25E-07 (7.93E-07) = & 6.85E-07 (1.32E-06) = & 4.05E-07 (4.82E-07) = & 7.64E-07 (1.27E-06) = \\
    F7    & 7.33E+01 (8.29E+00) & 7.45E+01 (8.12E+00) = & 7.21E+01 (7.05E+00) = & 7.42E+01 (9.08E+00) = & 7.35E+01 (6.44E+00) = & 7.14E+01 (7.06E+00) = & 7.17E+01 (6.10E+00) = & 7.08E+01 (5.93E+00) = \\
    F8    & 1.57E+01 (4.72E+00) & 1.61E+01 (4.59E+00) = & 1.57E+01 (5.85E+00) = & 1.74E+01 (5.91E+00) = & 1.53E+01 (4.67E+00) = & 1.43E+01 (4.37E+00) = & 1.46E+01 (4.78E+00) = & 1.54E+01 (3.97E+00) = \\
    F9    & 4.02E-14 (5.50E-14) & 2.68E-14 (4.88E-14) = & 4.02E-14 (5.50E-14) = & 3.58E-14 (5.34E-14) = & 3.58E-14 (5.34E-14) = & 2.24E-14 (4.57E-14) = & 1.79E-14 (4.19E-14) = & 2.68E-14 (4.88E-14) = \\
    F10   & 4.19E+03 (6.37E+02) & 4.66E+03 (6.18E+02) - & 4.55E+03 (6.74E+02) - & 4.34E+03 (5.03E+02) = & 3.92E+03 (6.17E+02) + & 3.81E+03 (6.12E+02) + & 3.94E+03 (5.35E+02) + & 3.74E+03 (6.22E+02) + \\
    F11   & 1.84E+01 (3.32E+00) & 1.91E+01 (4.69E+00) = & 2.01E+01 (4.16E+00) - & 1.83E+01 (3.70E+00) = & 1.82E+01 (3.70E+00) = & 1.96E+01 (3.99E+00) = & 1.99E+01 (4.17E+00) = & 2.20E+01 (3.91E+00) - \\
    F12   & 1.41E+03 (3.66E+02) & 1.08E+03 (3.01E+02) + & 1.26E+03 (4.28E+02) + & 1.45E+03 (4.26E+02) = & 1.44E+03 (3.63E+02) = & 1.54E+03 (3.89E+02) = & 1.57E+03 (3.42E+02) - & 1.32E+03 (3.84E+02) = \\
    F13   & 2.49E+01 (1.69E+01) & 2.75E+01 (2.14E+01) = & 3.04E+01 (1.91E+01) = & 2.58E+01 (1.63E+01) = & 2.86E+01 (2.39E+01) = & 2.47E+01 (1.85E+01) = & 2.97E+01 (2.10E+01) = & 3.26E+01 (1.87E+01) - \\
    F14   & 2.42E+01 (2.59E+00) & 2.37E+01 (2.06E+00) = & 2.45E+01 (1.95E+00) = & 2.37E+01 (2.22E+00) = & 2.36E+01 (1.95E+00) = & 2.36E+01 (1.93E+00) = & 2.36E+01 (2.21E+00) = & 2.34E+01 (1.99E+00) = \\
    F15   & 1.80E+01 (1.55E+00) & 1.86E+01 (1.92E+00) = & 1.83E+01 (1.73E+00) = & 1.80E+01 (1.99E+00) = & 1.93E+01 (1.93E+00) - & 1.97E+01 (2.06E+00) - & 2.06E+01 (2.30E+00) - & 2.06E+01 (1.84E+00) - \\
    F16   & 3.58E+02 (1.49E+02) & 3.37E+02 (1.49E+02) = & 3.48E+02 (1.26E+02) = & 3.10E+02 (1.28E+02) = & 3.22E+02 (1.42E+02) = & 3.03E+02 (1.14E+02) + & 3.16E+02 (1.28E+02) = & 3.20E+02 (9.99E+01) = \\
    F17   & 2.59E+02 (1.06E+02) & 2.82E+02 (1.39E+02) = & 2.92E+02 (1.30E+02) = & 2.53E+02 (7.98E+01) = & 2.26E+02 (9.65E+01) = & 2.21E+02 (8.00E+01) = & 2.18E+02 (8.77E+01) + & 2.24E+02 (7.52E+01) = \\
    F18   & 2.27E+01 (1.32E+00) & 2.29E+01 (1.49E+00) = & 2.30E+01 (1.37E+00) = & 2.28E+01 (1.40E+00) = & 2.29E+01 (1.71E+00) = & 2.27E+01 (1.40E+00) = & 2.32E+01 (1.36E+00) = & 2.25E+01 (1.04E+00) = \\
    F19   & 1.04E+01 (2.15E+00) & 9.22E+00 (2.37E+00) + & 9.61E+00 (1.80E+00) = & 1.02E+01 (1.96E+00) = & 1.05E+01 (2.00E+00) = & 1.15E+01 (2.67E+00) = & 1.08E+01 (2.42E+00) = & 1.06E+01 (2.72E+00) = \\
    F20   & 1.18E+02 (3.65E+01) & 1.88E+02 (1.17E+02) - & 1.53E+02 (6.84E+01) - & 1.34E+02 (7.27E+01) = & 1.06E+02 (3.39E+01) + & 1.14E+02 (3.91E+01) = & 1.11E+02 (3.58E+01) = & 1.01E+02 (1.78E+01) + \\
    F21   & 2.15E+02 (5.23E+00) & 2.16E+02 (4.81E+00) = & 2.17E+02 (5.51E+00) = & 2.16E+02 (4.97E+00) = & 2.15E+02 (4.61E+00) = & 2.16E+02 (4.24E+00) = & 2.15E+02 (4.74E+00) = & 2.14E+02 (4.27E+00) = \\
    F22   & 1.65E+03 (2.15E+03) & 1.73E+03 (2.34E+03) = & 1.61E+03 (2.18E+03) = & 1.74E+03 (2.28E+03) = & 1.80E+03 (2.24E+03) = & 1.37E+03 (2.00E+03) = & 1.24E+03 (1.91E+03) = & 1.08E+03 (1.83E+03) = \\
    F23   & 4.34E+02 (6.62E+00) & 4.34E+02 (7.70E+00) = & 4.33E+02 (7.11E+00) = & 4.31E+02 (8.02E+00) = & 4.31E+02 (6.21E+00) + & 4.31E+02 (6.96E+00) = & 4.30E+02 (6.45E+00) + & 4.31E+02 (7.17E+00) = \\
    F24   & 5.09E+02 (3.67E+00) & 5.09E+02 (4.60E+00) = & 5.09E+02 (3.73E+00) = & 5.09E+02 (4.74E+00) = & 5.08E+02 (3.49E+00) = & 5.09E+02 (3.75E+00) = & 5.10E+02 (3.62E+00) = & 5.09E+02 (3.63E+00) = \\
    F25   & 4.80E+02 (1.20E+00) & 4.80E+02 (1.83E+00) - & 4.80E+02 (1.48E+00) - & 4.79E+02 (9.18E-01) = & 4.79E+02 (1.10E+00) = & 4.79E+02 (6.66E-01) = & 4.80E+02 (7.00E-01) - & 4.80E+02 (7.84E-01) - \\
    F26   & 1.12E+03 (4.92E+01) & 1.14E+03 (5.97E+01) = & 1.13E+03 (5.17E+01) = & 1.13E+03 (4.77E+01) = & 1.13E+03 (5.25E+01) = & 1.13E+03 (4.12E+01) = & 1.13E+03 (6.13E+01) = & 1.13E+03 (5.02E+01) = \\
    F27   & 4.79E+02 (6.08E+00) & 5.00E+02 (0.00E+00) - & 5.00E+02 (0.00E+00) - & 4.86E+02 (1.02E+01) - & 5.03E+02 (1.17E+01) - & 5.08E+02 (9.27E+00) - & 5.11E+02 (1.19E+01) - & 5.10E+02 (8.80E+00) - \\
    F28   & 4.52E+02 (2.44E-01) & 4.99E+02 (5.23E-01) - & 4.97E+02 (1.89E+00) - & 4.52E+02 (5.05E-01) - & 4.52E+02 (3.25E-01) = & 4.52E+02 (3.11E-01) = & 4.53E+02 (4.40E-01) - & 4.54E+02 (6.70E+00) - \\
    F29   & 3.15E+02 (2.82E+01) & 3.03E+02 (1.85E+01) + & 2.92E+02 (1.35E+01) + & 2.98E+02 (1.70E+01) + & 3.75E+02 (1.95E+01) - & 3.70E+02 (1.78E+01) - & 3.73E+02 (1.70E+01) - & 3.71E+02 (2.10E+01) - \\
    F30   & 4.06E+03 (4.03E+03) & 5.75E+02 (1.42E+02) + & 1.05E+03 (3.43E+02) + & 3.93E+03 (2.93E+03) = & 4.08E+05 (5.34E+04) - & 5.54E+05 (5.52E+04) - & 5.89E+05 (3.07E+04) - & 5.98E+05 (2.60E+04) - \\
    \midrule
    +/=/- &       & 4/20/6 & 4/19/7 & 1/25/4 & 5/21/4 & 3/23/4 & 3/20/7 & 3/19/8 \\
    \bottomrule
    \end{tabular}%
  \label{tab:Alpha}%
\footnotesize
\\The symbols ``+/=/-" indicate that the corresponding algorithm performed significantly better ($+$), not significantly better or worse ($=$), or significantly worse ($-$) compared to iLSHADE-RSP with $\alpha = 1.0$ using the Wilcoxon rank-sum test with $\alpha = 0.05$ significance level.
\end{table*}%

\subsection{Analysis of iLSHADE-RSP}

\subsubsection{Parameter Values for Jumping Rate}
As we mentioned earlier, the proposed algorithm alternately applies one of the two recombination operators according to the jumping rate $p_{j}$. The jumping rate determines the additional exploration of the proposed algorithm. If the jumping rate is too high, the modified recombination operator is applied too often, and thus, the proposed algorithm might not be beneficial from existing candidate solutions. On the other hand, if the jumping rate is too low, the modified recombination operator is applied too rarely, and thus, the additional exploration of the proposed algorithm might be negligible. We carried out experiments to find out appropriate parameter values for the jumping rate. Table \ref{tab:JumpingRate} shows the means and standard deviations of the FEVs of the proposed algorithm with different parameter values for the jumping rate. As can be seen from the table, the parameter values $p_{j} \in [0.15, 0.35]$ can lead to satisfactory results.

\subsubsection{L\'{e}vy Perturbation}
As we mentioned earlier, the proposed algorithm employs a modified recombination operator, which calculates a perturbation of a target vector with the Cauchy distribution. The Cauchy distribution is a special case of the L\'{e}vy $\alpha$-stable distribution. Therefore, it is interesting to test the other cases of the L\'{e}vy $\alpha$-stable distribution for the modified recombination operator. The L\'{e}vy $\alpha$-stable distribution can be defined by the characteristic function, derived by L\'{e}vy \cite{levy1934integrales} and Hall \cite{hall1981comedy} as follows.

\begin{equation}
log\o(t) = \left\{ \begin{array}{ll}
- c^{\alpha} |t| \{ 1 - i \beta sign(t) tan\frac{\pi \alpha}{2} \} + i \mu t & \textrm{if $\alpha \not= 1$} \\
- c |t| \{ 1 + i \beta sign(t) \frac{2}{\pi} log |t| \} + i \mu t & \textrm{if $\alpha = 1$}
\end{array} \right.
\end{equation}

\noindent
where

\begin{equation}
sign(t) = \left\{ \begin{array}{lll}
1 & \textrm{if $t > 0$} \\
0 & \textrm{if $t = 0$} \\
-1 & \textrm{if $t < 0$}
\end{array} \right.
\end{equation}

The L\'{e}vy $\alpha$-stable distribution denoted by $S_{\alpha}(\beta, c, \mu)$ has four parameters: the stability parameter $\alpha \in (0, 2]$, the skewness parameter $\beta \in [-1, 1]$, the scale parameter $c \in (0, \infty)$, and the location parameter $\mu \in (-\infty, \infty)$. The L\'{e}vy $\alpha$-stable distribution has three special cases as follows.

\begin{itemize}
    \item The Gaussian distribution: $S_{2}(\beta, 2c^{2}, \mu)$.
    \item The Cauchy distribution: $S_{1}(0, c, \mu)$.
    \item The L\'{e}vy distribution: $S_{0.5}(1, c, \mu)$.
\end{itemize}

The modified recombination operator, which calculates a perturbation of a target vector with the L\'{e}vy $\alpha$-stable distribution can be defined as follows.

\begin{equation}
u_{i,g}^{j} = \left\{ \begin{array}{ll}
x_{i,g}^{j} + F_{w} \cdot (x_{pbest,g}^{j} - x_{i,g}^{j}) & \textrm{if $rand_{i}^{j} < CR$ or $j = j_{rand}$} \\
\,\,\,\,\,\,\,\,\, + F \cdot (x_{pr_{1},g}^{j} - \tilde{x}_{pr_{2},g}^{j}) & \\
S_{\alpha}(0, 0.1, x_{i,g}^{j}) & \textrm{otherwise}
\end{array} \right.
\label{modifiedRecombination}
\end{equation}

We used Chambers-Mallows-Stuck method \cite{chambers1976method, weron1996chambers, weron2010correction, weron2004computationally} to generate L\'{e}vy $\alpha$-stable random numbers as follows.

\begin{quote}
\textbf{Step 1.} Generate a random number from the uniform distribution $V \in [-\frac{\pi}{2},\frac{\pi}{2}]$ and a random number from the exponential distribution $W$ with mean 1.
\end{quote}

\begin{quote}
\textbf{Step 2.} If $\alpha \not= 1$ then:

\begin{equation}
X = S_{\alpha, \beta} \cdot \frac{sin \{ \alpha ( V + B_{\alpha, \beta} ) \}}{\{ cos ( V ) \}^{\frac{1}{\alpha}}} \cdot \left[ \frac{cos \{ V - \alpha ( V + B_{\alpha, \beta} ) \}}{W} \right]^{\frac{( 1 - \alpha )}{\alpha}}
\end{equation}

\noindent
where

\begin{equation}
B_{\alpha, \beta} = \frac{arctan( \beta tan \frac{\pi \alpha}{2} )}{\alpha}
\end{equation}
\begin{equation}
S_{\alpha, \beta} = \Big\{ 1 + \beta^{2} tan^{2} \left( \frac{\pi \alpha}{2} \right) \Big\}^{\frac{1}{(2 \alpha)}}
\end{equation}
\end{quote}

\begin{quote}
\textbf{Step 3.} If $\alpha = 1$ then:

\begin{equation}
X = \frac{2}{\pi} \cdot \left( \frac{\pi}{2} + \beta V \right) tan V - \frac{2}{\pi} \cdot \beta ln \left( \frac{\frac{\pi}{2} W cos V}{\frac{\pi}{2} + \beta V} \right)
\end{equation}
\end{quote}

\begin{quote}
\textbf{Step 4.} If $X \sim S_{\alpha}(\beta, 1, 0)$ then:

\begin{equation}
Y = \left\{ \begin{array}{ll}
c X + \mu & \textrm{if $\alpha \not= 1$} \\
c X + \frac{2}{\pi} \beta c log c + \mu & \textrm{if $\alpha = 1$}
\end{array} \right.
\end{equation}

is $S_{\alpha}(\beta, c, \mu)$.
\end{quote}

Table \ref{tab:Alpha} shows the means and standard deviations of the FEVs of the proposed algorithm with different stability parameters for the L\'{e}vy $\alpha$-stable distribution with the jumping rate $p_{j} = 0.2$. For simplicity's sake, we only considered symmetric distributions. The L\'{e}vy $\alpha$-stable distribution has a short and wide PDF if the stability parameter is high, while a tall and narrow PDF if the stability parameter is low. As can be seen from the table, the parameter values $\alpha = [1, 1.5]$ can lead to satisfactory results. In other words, the modified recombination operator with from the Cauchy distribution ($\alpha = 1$) to the L\'{e}vy $\alpha$-stable distribution ($\alpha = 1.5$) works best for the proposed algorithm. We chose the Cauchy distribution for the modified recombination operator because generating Cauchy random numbers is much easier than generating L\'{e}vy $\alpha$-stable random numbers.



\section{Conclusion}
\label{sec:Conclusion}
Differential evolution is a popular evolutionary algorithm for multidimensional real-valued functions. Like other evolutionary algorithms, it is important for differential evolution to establish a balance between exploration and exploitation to be successful. Recently, a state-of-the-art DE variant called LSHADE-RSP was proposed. Although it has shown excellent performance, the greediness of LSHADE-RSP may cause premature convergence in which all the candidate solutions fall into the local optimum of an optimization problem and cannot escape from there.

To mitigate the problem, we have devised a modified recombination operator for LSHADE-RSP, which perturbs a target vector with the Cauchy distribution. Therefore, the modified recombination operator can increase the probability of finding an optimal solution by adopting the long-tailed property of the Cauchy distribution. We called the resulting algorithm iLSHADE-RSP, which alternately applies the original and modified recombination operators according to a jumping rate.

The proposed algorithm has been tested on the CEC 2017 test suite in 10, 30, 50, and 100 dimensions. Our experimental results verify that the improved LSHADE-RSP significantly outperformed not only its predecessor LSHADE-RSP but also several cutting-edge DE variants in terms of convergence speed and solution accuracy. In particular, the proposed algorithm performs better than the comparison algorithms on more complicated optimization problems, such as expanded multimodal test functions and hybrid composition test functions, in all the dimensions.

Possible directions for future work include 1) testing the proposed algorithm for constrained optimization problems; 2) testing the proposed algorithm for large-scale optimization problems; 3) applying the proposed algorithm to various real-world problems.




\section*{Acknowledgement}

This work was supported by the National Research Foundation of Korea(NRF) grant funded by the Korea government(MSIT) (No. NRF-2017R1C1B2012752). The correspondence should be addressed to Dr. Chang Wook Ahn.

\bibliographystyle{elsarticle-num} 
\bibliography{refs}





\end{document}